\documentclass[10pt, a4paper]{article}

\usepackage[final]{lrec2026} 

\usepackage{amsmath,amssymb,amsfonts}
\usepackage{algorithmic}
\usepackage{graphicx}
\usepackage{textcomp}
\usepackage{multirow}
\usepackage{comment}
\usepackage{subcaption}
\usepackage{fontawesome}
\usepackage[misc]{ifsym}
\newcommand{\fstar}{\textsuperscript{\fontsize{6pt}{6pt}\selectfont \faStarO}}
\newcommand{\corres}{\textsuperscript{\fontsize{7pt}{6pt}\selectfont \Letter}}

\usepackage{booktabs}
\usepackage{tabularx}
\usepackage{array} 

\newcolumntype{Y}{>{\raggedright\arraybackslash}X}

\title{CEFR-Annotated WordNet: LLM-Based Proficiency-Guided Semantic Database for Language Learning}

\name{
{\bf \large Masato Kikuchi\textsuperscript{$\diamondsuit$}\corres\thanks{\Letter \quad Corresponding author.}},
{\bf \large Masatsugu Ono\textsuperscript{$\heartsuit$}},
{\bf \large Toshioki Soga\fstar},\\
{\bf \large Tetsu Tanabe\textsuperscript{$\clubsuit$}},
{\bf \large Tadachika Ozono\textsuperscript{$\diamondsuit$}}
}

\address{
\textsuperscript{$\diamondsuit$}Nagoya Institute of Technology,
\textsuperscript{$\heartsuit$}Kitami Institute of Technology, \\
\textsuperscript{\faStarO}Chitose Institute of Science and Technology,
\textsuperscript{$\clubsuit$}Hokkaido University \\
\{kikuchi, ozono\}@nitech.ac.jp,
onomasa@mail.kitami-it.ac.jp, \\
t-soga@photon.chitose.ac.jp,
ttanabe@iic.hokudai.ac.jp
}

\abstract{
Although WordNet is a valuable resource because of its structured semantic networks and extensive vocabulary, its fine-grained sense distinctions can be challenging for second-language learners.
To address this issue, we developed a version of WordNet annotated with the Common European Framework of Reference for Languages (CEFR), integrating its semantic networks with language-proficiency levels.
We automated this process using a large language model to measure the semantic similarity between sense definitions in WordNet and entries in the English Vocabulary Profile Online.
To validate our approach, we constructed a large-scale corpus containing both sense and CEFR-level information from the annotated WordNet and used it to develop contextual lexical classifiers.
Our experiments demonstrate that models fine-tuned on this corpus perform comparably to those fine-tuned on gold-standard annotations.
Furthermore, by combining this corpus with the gold-standard data, we developed a practical classifier that achieves a Macro-F1 score of 0.81. This result provides indirect evidence that the transferred labels are largely consistent with the gold-standard levels.
The annotated WordNet, corpus, and classifiers are publicly available to help bridge the gap between natural language processing and language education, thereby facilitating more effective and efficient language learning.
 \\ \newline \Keywords{WordNet, CEFR Level, Language Learning, Word Sense, Corpus} }

\begin{document}

\maketitleabstract

\section{Introduction}
\label{sec:introduction}

WordNet~\citelanguageresource{Fellbaum:98} is a large-scale English lexical database that organizes approximately 155,000 words and 207,000 senses of nouns, verbs, adjectives, and adverbs into hierarchical semantic networks. 
It groups semantically similar words and links senses through relations such as hypernymy, hyponymy, synonymy, and antonymy. 
Because WordNet and its construction software are publicly available, they can be readily integrated into AI applications. 
Consequently, WordNet underpins a broad range of natural language processing (NLP) technologies---including machine translation~\cite{Moussallem:18}, semantic analysis~\cite{Moskvoretskii:2024}, and natural language generation~\cite{Vicente:2014}---owing to its accessible interface and well-structured networks. 
These technologies also support computer-assisted language learning (CALL) by facilitating vocabulary acquisition, reading comprehension, writing assistance, automated question generation, and automated assessment.

\begin{figure*}[tb]
  \centering
  \subcaptionbox{
    CEFR-Annotated WordNet.\label{fig:wordnet}
  }[0.48\linewidth]{
    \includegraphics[width=\linewidth]{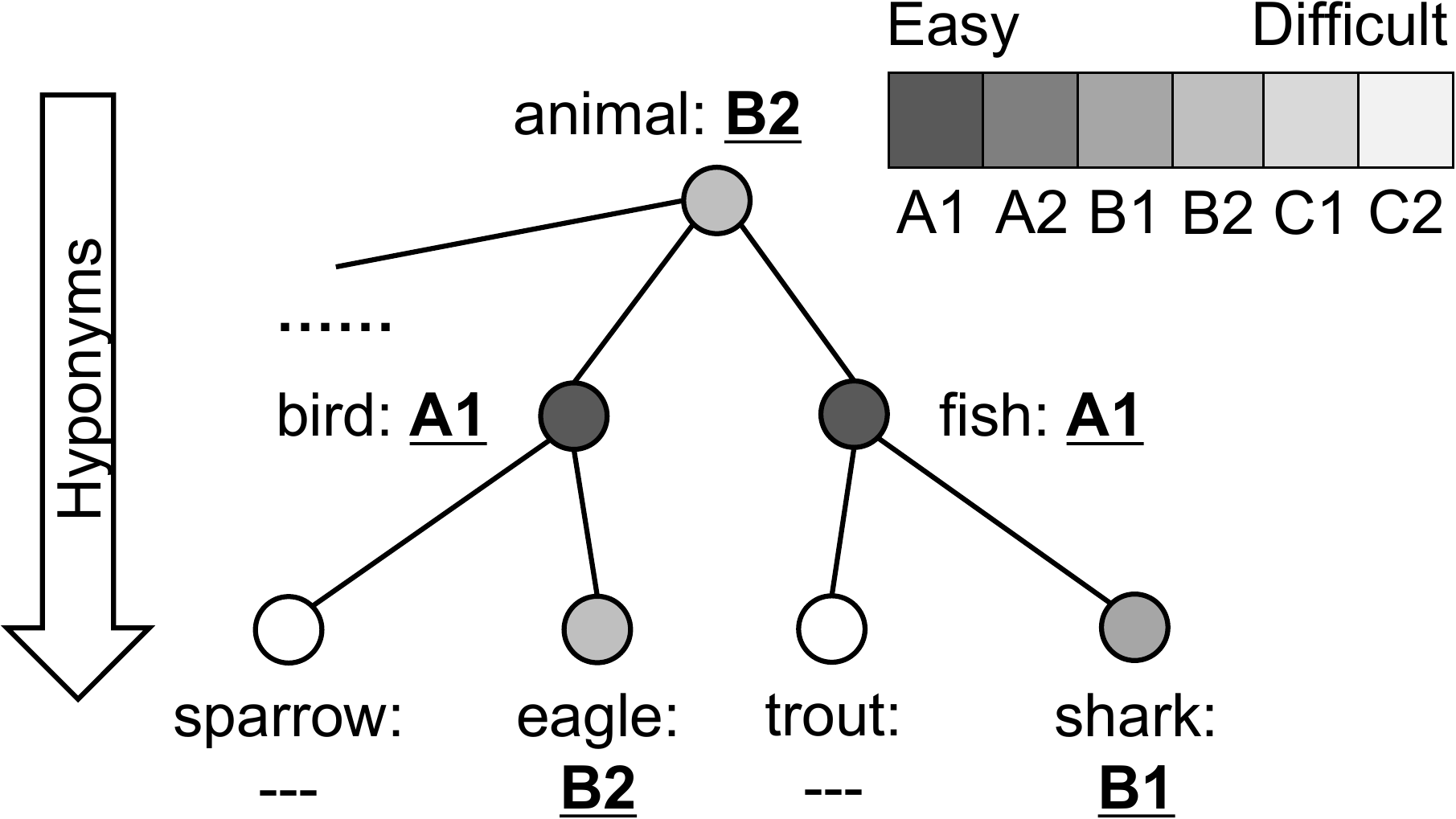}
  }
  \hfill
  \subcaptionbox{
    Contextual Lexical CEFR-Level Classifier.\label{fig:classifier}
  }[0.48\linewidth]{
    \includegraphics[width=\linewidth]{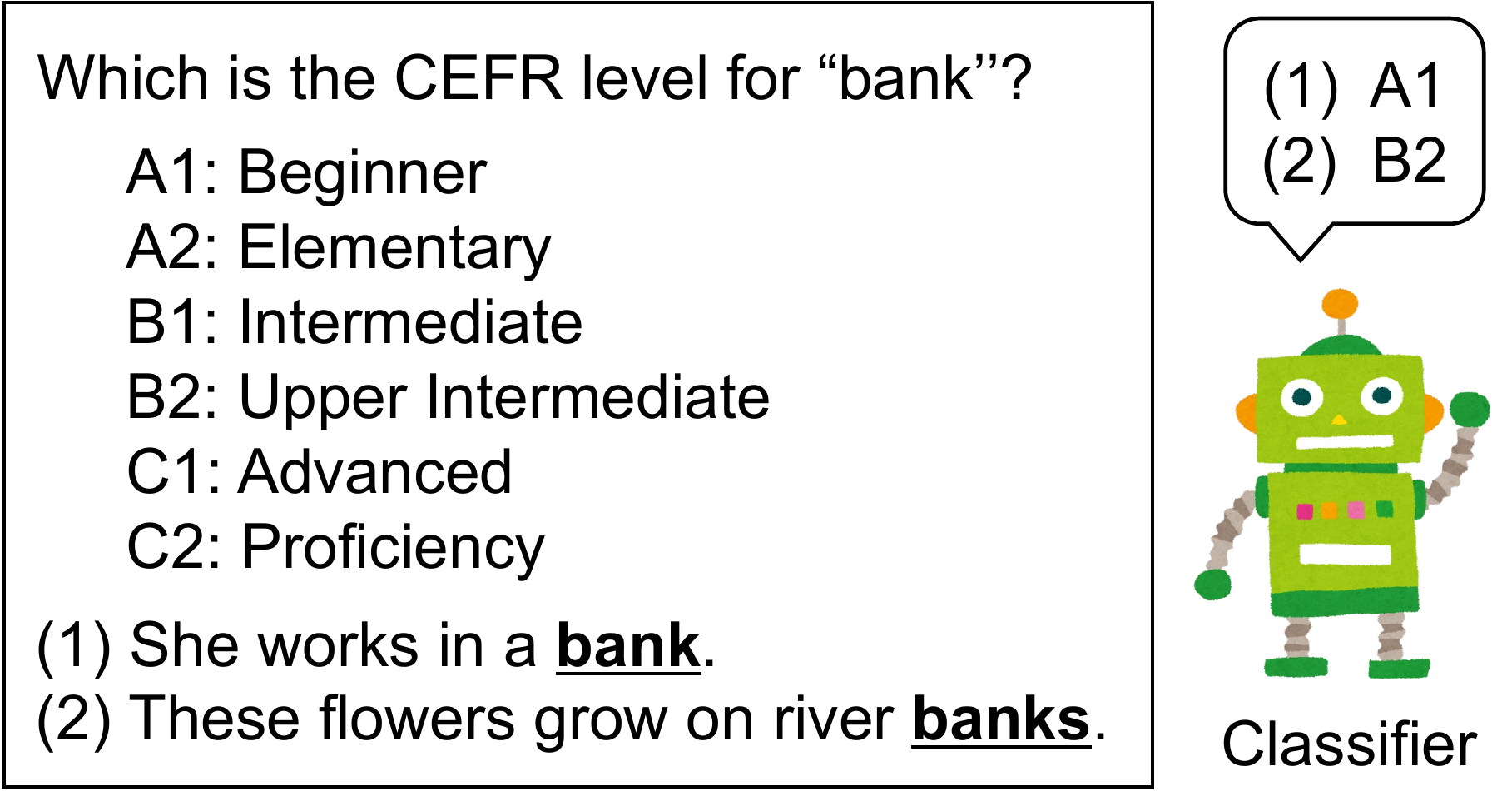}
  }
  \caption{Overview of the study. 
(a) Semantic network of hyponyms for ``animal'' in the CEFR-annotated WordNet.
(b) Contextual CEFR-level classification for the word ``bank.''
}
  \label{fig:overview}
\end{figure*}
Although leveraging semantic networks can enhance foreign-language learning~\cite{Kiritani:12}, WordNet was not designed for educational purposes and presents challenges for second-language (L2) learners. 
Key issues are its overly fine-grained sense distinctions and the large number of senses associated with many words. This requires learners to identify the appropriate sense for a given context and proficiency level, which increases their cognitive load. 
While this problem is widely discussed in NLP literature~\cite{Navigli:06}\citelanguageresource{Lacerra:20}, it has received limited attention in language education. 
Our goal was to develop a novel version of WordNet and leverage the resulting technologies and resources to enhance language-learning efficiency. The first step involves adapting WordNet for L2 learners and bridging the gap between NLP lexical resources and language education.
Previous work~\citelanguageresource{Kikuchi:24} clustered fine-grained WordNet sense definitions (glosses) using learner-oriented dictionaries. By contrast, this study integrates Common European Framework of Reference for Languages (CEFR) proficiency levels into WordNet, enabling the presentation of senses aligned with a learner's proficiency level.
To build large-scale, practical resources, we employ a simple large language model (LLM)-based method for efficient and accurate semantic annotation, reducing the time, labor, and cost associated with manual annotation. 
This automatic approach also ensures that the adapted WordNet can be flexibly scaled.

The CEFR is an international standard for describing language proficiency across six levels, namely, A1, A2, B1, B2, C1, and C2, ranging from basic to advanced. Each level is defined by ``can-do descriptors'' that specify expected communicative abilities.
We used an LLM to annotate WordNet senses with CEFR levels, thereby constructing a CEFR-annotated WordNet. 
As shown in Figure~\ref{fig:wordnet}, these annotations can be used in conjunction with semantic networks to help learners acquire vocabulary while considering relationships among words and learn basic and advanced paraphrases through synonyms.
The annotation pipeline involves three steps. First, we collect glosses for target words from WordNet and the English Vocabulary Profile (EVP) Online\footnote{\url{https://englishprofile.org/?menu=evp-online}}~\citelanguageresource{Capel:12}, which provides CEFR levels for individual senses. 
Next, the LLM computes the semantic similarity between the glosses from WordNet and the EVP. 
Finally, we assign CEFR levels to the corresponding WordNet senses based on these similarity scores.

Because our method for annotating WordNet senses with CEFR levels is automatic, it eliminates the need for labor-intensive manual work. 
However, automatic labels may contain errors, and WordNet does not provide gold-standard CEFR levels. Therefore, their reliability must be verified indirectly.
To address this issue, we built contextual CEFR-level classifiers that predict a sense's proficiency level from its usage, as shown in Figure~\ref{fig:classifier}.
These classifiers predict the CEFR level for a word sense based on its context, not just for the word itself. 
We evaluate the quality of our annotations by comparing classifiers trained on our data with those trained on the EVP gold-standard levels.
We also examine the effectiveness of LLMs for this task in few-shot and fine-tuning settings.

The contributions of this study can be summarized as follows:
\begin{itemize}
\item \textbf{CEFR-Annotated WordNet.} We developed a new resource by assigning CEFR proficiency levels to 10,644 WordNet senses corresponding to 5,645 lemmas, thereby linking the WordNet sense inventory with CEFR standards.
Our annotated WordNet covers approximately 80\% of all single-word senses in the EVP (8,289 out of 10,394).

\item \textbf{Prompt-Only LLM Annotation.} We introduce a novel, automated method that leverages the semantic understanding of LLMs to assign CEFR levels to word senses. 
This is achieved by measuring semantic similarity between WordNet glosses and EVP entries.
The method, implemented entirely through prompting, is simple, reproducible, and substantially less costly than manual annotation. We also provide indirect evidence that manual annotation tasks based on semantic matching can be automated with high accuracy.

\item \textbf{SemCor-CEFR Corpus.} Using our annotated WordNet, we assigned CEFR levels to the word senses in SemCor 3.0\footnote{\url{http://lcl.uniroma1.it/wsdeval/training-data}}~\citelanguageresource{Miller:93}, a widely used sense-annotated corpus. This process resulted in a large-scale corpus containing more than 110,000 sense and level annotations across over 5,500 WordNet senses.
Because modern NLP relies on large corpora for advanced training and analysis, this resource represents a valuable contribution to both NLP and educational technology research.

\item \textbf{Contextual Lexical CEFR-Level Classifiers.} We demonstrate the validity of our CEFR-level annotations by training a classifier on our corpus that performs comparably to one trained on gold-standard EVP data.
Additionally, by fine-tuning the LLM on both our annotated data and the gold-standard levels, we developed a practical classifier that achieves a Macro-F1 score of 0.81. 
Our analysis indicates that these classifiers can accurately predict CEFR levels across a broad range of contexts.
\end{itemize}

All resources developed in this study, including our WordNet, corpus, and classifier, are publicly available at \url{https://doi.org/10.5281/zenodo.17395388}.

\section{Related Work}

\subsection{WordNets for Language Learning}

As noted in the introduction, WordNet was not originally designed for educational use. 
To address this limitation, several learning-oriented WordNets have been developed for multiple languages~\citelanguageresource{Bosch:18}, and their application in language learning has been studied by many researchers~\cite{Gonzalez-Dios:19}. 
Some of these researchers have focused on visualizing word hierarchies and semantic relations to support learners~\cite{Sun:11,Kiritani:12,Gawde:24}, 
while others have tailored vocabulary, glosses, and usage examples to match learners' proficiency levels~\citelanguageresource{Redkar:18,Osenova:24}. 
However, most of these studies have relied on manually curated resources and paid limited attention to word-sense information.
By contrast, herein we introduce a novel approach that automatically annotates WordNet senses with proficiency levels. 
Our method can be integrated with existing techniques---such as semantic network visualization and multimodal WordNets~\cite{Marciniak:20}---to further enhance its utility in language-learning contexts.

\subsection{Lexical Complexity Prediction}

Lexical complexity prediction (LCP)~\cite{North:23,Shardlow:24} has recently attracted significant attention as a task for estimating word complexity from context. 
In this field, ``complexity''---which is related to the CEFR levels in our study---is typically predicted as either binary (e.g., simple/complex) or on a continuous scale. 
Our work is closely related to SemEval-2021 Task 1~\cite{Shardlow:21}, which adopts a similar classification setting. For this task, the organizers released the CompLex 2.0 dataset\footnote{\url{https://github.com/MMU-TDMLab/CompLex}}~\citelanguageresource{Shardlow:22}, in which words in context were rated by multiple annotators on a five-point Likert scale. 
The final scores are represented as a continuous value in $[0,1]$, computed as the mean of these ratings. 
These continuous values can capture finer, context-driven differences compared with ordinal labels.

However, our approach differs from that of LCP in several key ways. 

First, LCP is primarily designed as a precursor to lexical simplification~\cite{Paetzold:17}---which involves replacing complex words with simpler alternatives---rather than for explicitly presenting complexity information to L2 learners. 
Second, annotators of CompLex 2.0 were not provided with glosses; as a result, identical senses may receive different scores across contexts.
Third, the dataset is limited to 9,000 nouns and excludes other parts of speech (PoS).

By contrast, our method assigns a CEFR level to each word sense in accordance with an international proficiency standard.
We extend this annotation to more than 110,000 instances of nouns, verbs, adjectives, and adverbs in the large-scale SemCor corpus, making our resource more than ten times larger than CompLex 2.0. 
In Section~\ref{sec:correlation_analysis}, we analyze the correlation between the CompLex 2.0 complexity scores and the CEFR levels predicted by our models.

\subsection{CEFR-Based Educational Technology}

The CEFR is a foundational standard in educational technology and is widely applied in the automatic assessment of short sentences~\cite{Tack:17,Uchida:24}, teaching materials~\cite{Pilan:16}, writing skills~\cite{Kerz:21,Schmalz:21}, and learner proficiency~\cite{Gaillat:22}. 
The interaction between LLMs and the CEFR has been the focus of recent studies, which have explored how well these models understand proficiency levels~\cite{Benedetto:25} and how to control the difficulty of the vocabulary and sentences they generate~\cite{Alfter:24,Malik:24,Barayan:25}. 
These studies, together with the development of numerous CEFR-aligned lexical datasets, underscore the central role of the CEFR in the field.

For example, the CEFRLex project provides machine-readable lexical resources with word-level frequency counts by CEFR level~\cite{Pintard:20} for English and other languages\footnote{Only the Dutch resource NT2Lex provides frequency information at the word-sense level.}~\citelanguageresource{Durlich:18,Francois:14,Tack:18,Francois:16,Volodina:16}.
However, it does not assign a unique CEFR level to each sense, making it unsuitable for tasks requiring sense-specific proficiency annotations. 
The Sense Complexity Dataset (SeCoDa)~\citelanguageresource{Strohmaier:20} provides sense-in-context CEFR annotations, but its sense labels are not aligned with those of WordNet. In addition, its small size, 1,432 words, limits its applicability within WordNet's semantic framework. 
Our work addresses these gaps by annotating more than 110,000 word instances with sense-specific CEFR levels, thereby substantially expanding the available data on lexical difficulty.

\begin{figure*}[tb]
\centering
\includegraphics[width=\linewidth]{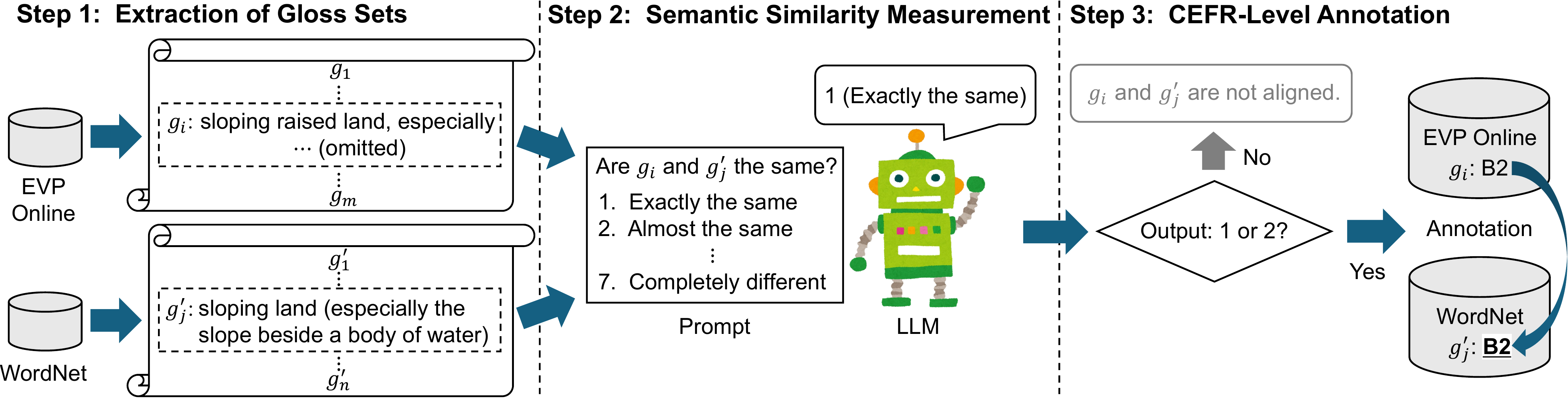}
\caption{CEFR-level annotation process for a WordNet gloss ($g'_j$), illustrated with the noun ``bank.''}
\label{fig:annotation}
\end{figure*}
Despite progress in LCP, few studies have examined the classification of vocabulary into CEFR levels based on context.
\citet{Aleksandrova:23} proposed ME6 Contextual, a BERT~\cite{Devlin:19}-based classifier that, like our models, is trained on a CEFR-annotated corpus to directly predict a word's level from its context. This direct prediction approach enables the model to classify words not seen during training.
By contrast, \citet{Banno:25} introduced an indirect-prediction method in which an LLM selects the appropriate EVP sense for a word in context and then maps it to a CEFR level. The performance of this approach depends on the quality of cues provided by the data source.
To isolate the effect of different data sources on classification performance, we reimplemented ME6 Contextual as a baseline for our LLM-based classifiers.

\section{Existing Resources}
\subsection{EVP Online}

The EVP Online\footnotemark[1], developed by Cambridge University Press, provides CEFR levels for single words, phrasal verbs, phrases, and idioms.
Each entry includes a PoS tag, a gloss, and both dictionary and learner examples.
A key feature of the EVP is its sense-level CEFR annotation, which assigns a proficiency level to each sense. 
This level of granularity is beneficial for both general language education and the development of CALL systems.
For this study, we used single-word entries from the American English subset of the EVP, including their CEFR levels, PoS tags, glosses, and example sentences. Unlike a previous study~\cite{Aleksandrova:23}, which also included multiword expressions (MWEs), our work focuses exclusively on single words because WordNet contains very few MWEs.

\subsection{SemCor Corpus}
\label{sec:semcor}

The SemCor 3.0 corpus is one of the most widely used sense-annotated corpora in NLP.
It contains 226,040 sense annotations across 352 documents from the Brown Corpus.
Each sense is tagged with a WordNet identifier, linking it to glosses, usage examples, and semantic relations such as hypernyms, hyponyms, synonyms, and antonyms. 
Moreover, its machine-readable format facilitates integration into NLP and CALL systems.
However, this corpus inherits the limitations of WordNet discussed in the introduction. In particular, the absence of learner-oriented indicators, such as sense complexity or CEFR levels, limits its usefulness for educational applications.
To address this issue, as described in Section~\ref{sec:settings}, we use our CEFR-annotated WordNet to enhance the original SemCor corpus, creating a new resource annotated with both senses and CEFR levels.

\section{CEFR-Annotated WordNet}
\label{sec:our_wordnet}

To create a WordNet oriented toward L2 learners, we annotated its senses with CEFR levels by aligning them with glosses from the EVP Online. 
The process, illustrated in Figure~\ref{fig:annotation} for the WordNet gloss $g'_j$ of $\langle$word, PoS$\rangle=\langle$bank, noun$\rangle$, comprises three steps:

\begin{table*}[tb]
\renewcommand{\arraystretch}{1.1}
\centering
\begin{tabular}{lrrrrrrrr}
\toprule
\multicolumn{1}{ l }{\multirow{2}{*}{PoS}} & \multicolumn{6}{ c }{CEFR Levels} & \multicolumn{1}{ c }{\multirow{2}{*}{Total}} & \multicolumn{1}{ c }{\multirow{2}{*}{Share (\%)}} \\ \cmidrule{2-7}
& A1 & A2 & B1 & B2 & C1 & C2 & & \\
\midrule
Noun       & 310 & 626 & 1,021 & 1,426 & 652 & 853 & 4,888 & 44.46 \\
Verb       & 213 & 263 &  701 &  948 & 443 & 595 & 3,163 & 28.77 \\
Adjective  & 104 & 200 &  435 &  646 & 423 & 519 & 2,327 & 21.16 \\
Adverb     &  40 &  94 &  127 &  201 &  92 &  63 & 617 & 5.61 \\
\midrule
Total      & 667 & 1,183 & 2,284 & 3,221 & 1,610 & 2,030 & 10,995 & 100.00 \\
Share (\%) & 6.07 & 10.76 & 20.77 & 29.30 & 14.64 & 18.46 & 100.00 & \\
\bottomrule
\end{tabular}
\caption{Distribution of senses in the CEFR-annotated WordNet by PoS and CEFR level. Share (\%) indicates proportions by PoS (right) and level (bottom). Note that some senses received multiple levels because of differences in gloss granularity between the resources (see Appendix~\ref{sec:multi_labeled_sense_keys} for details).
}
\label{tab:summary}
\end{table*}
\paragraph{Step 1: Extraction of Gloss Sets.}
For each word and PoS pair, such as $\langle$bank, noun$\rangle$, we extract all corresponding glosses from both the EVP Online and WordNet.
Let the set of $m$ glosses from the EVP Online be $\{g_1, g_2, \ldots, g_m\}$, and that of $n$ glosses from WordNet be $\{g'_1, g'_2, \ldots, g'_n\}$.
In the next step, we compare the $i$-th gloss, $g_i$, from the EVP Online with the $j$-th gloss, $g'_j$, from WordNet. 
As shown in Figure~\ref{fig:annotation}, both example glosses refer to sloping land.

\paragraph{Step 2: Semantic Similarity Measurement.}
To measure the semantic similarity between $g_i$ and $g'_j$, we used an LLM (GPT-4o, checkpoint \texttt{gpt-4o-2024-08-06}).
 The prompt is provided in Appendix~\ref{sec:appendix_a}.
Because glosses from different resources often vary in granularity and may not align perfectly, a binary alignment judgment of same or different would be overly restrictive.
Therefore, we instructed the LLM to rate similarity on a seven-point scale, where a lower score indicates higher similarity. 
In the example shown in Figure~\ref{fig:annotation}, the LLM returns a score of 1, indicating that the two glosses have identical meanings.

\paragraph{Step 3: CEFR-Level Annotation.}
If the LLM returns a score of 1 or 2---indicating that $g_i$ and $g'_j$ have ``exactly the same'' or ``almost the same'' meaning---we consider the glosses semantically aligned.
The CEFR level associated with $g_i$ is then transferred to $g'_j$. Otherwise, that is, if the output is $\ge 3$, the senses are considered mismatched and no annotation is assigned.
In the example, the score of 1 results in the WordNet sense being assigned the B2 level from the corresponding EVP sense.
In this study, we adopt a threshold of $\le 2$ (scores 1--2) to balance accuracy and coverage.
Restricting alignments to score 1 yields higher-confidence transfers but reduces coverage, whereas allowing score 3 increases coverage but tends to introduce false alignments, mainly because of partial semantic overlap and mismatched gloss granularity across resources. Therefore, we adopt $\le 2$ as a conservative compromise.

We applied this procedure to all gloss pairs for every $\langle$word, PoS$\rangle$ entry found in both the EVP Online and WordNet.
For instance, the set of all possible gloss pairs of $\langle$bank, noun$\rangle$ is
\begin{multline*}
\{ (g,\ g') \mid g \in \{ g_1,\ g_2,\ \dots,\ g_m \},\\ g' \in \{ g'_1,\ g'_2,\ \dots,\ g'_n \} \},
\end{multline*}
whose size is $m \times n$.
This exhaustive process yielded 10,995 CEFR-level annotations for 10,644 WordNet senses across 5,645 lemmas.
Table~\ref{tab:summary} lists the distribution of these annotations. Nouns constitute the largest share (4,888; 44.46\%), followed by verbs (3,163; 28.77\%) and adjectives (2,327; 21.16\%), with adverbs comparatively scarce (617; 5.61\%).
The distribution across CEFR levels is concentrated in the intermediate range, with B2 (29.30\%) and B1 (20.77\%) together accounting for half of all annotations. The beginner levels (A1–A2) and advanced levels (C1–C2) represent 16.83\% and 33.10\%, respectively.
Because the granularity of glosses differs between the two resources, a single WordNet sense may align with multiple EVP glosses, which can result in a sense being assigned multiple CEFR levels (see Appendix~\ref{sec:multi_labeled_sense_keys} for details).
This automated procedure is generalizable and can be applied to other dictionaries or lexical databases that provide glosses. Moreover, because the pipeline relies solely on semantic similarity between annotations, it is transferable to lexical resources in other languages. However, as the process is fully automated, evaluating the accuracy of the resulting annotations is essential.

\section{Experiments}
\label{sec:experiment}

To verify the accuracy of our CEFR-level annotations, we built and evaluated several contextual lexical CEFR-level classifiers (Figure~\ref{fig:classifier}).
The goal was to assess how well models trained on our automatically annotated data could predict gold-standard CEFR levels. 
We also trained several LLM-based classifiers to evaluate their effectiveness for this task.
In this study, we did not conduct a manual spot-check of the transferred sense-level labels.
Although expert validation would strengthen the reliability of the resource, rigorous CEFR-level judgment typically requires carefully designed protocols and multiple trained raters. We therefore leave such manual evaluation to future work.

\subsection{Datasets and Experimental Settings}
\label{sec:settings}

To train our classifiers, we required a corpus with CEFR-level annotations for words in context. Because the original SemCor corpus does not include this information, we created the ``SemCor-CEFR corpus'' by assigning CEFR levels to its senses using our annotated WordNet. 
Table~\ref{tab:distribution} summarizes the word distributions in the EVP Online examples, combining dictionary and learner examples, and in the SemCor-CEFR corpus. Although our corpus contains fewer word types (\# types) than the EVP examples, it includes substantially more word instances (\# words) and reflects a more natural and imbalanced distribution of proficiency levels.
For our experiments, we used 10\% of the EVP examples as the test set.
For training and validation, depending on the setting, we used (i) the remaining 90\% of the EVP data, (ii) the SemCor-CEFR corpus, or (iii) a combination of both.
The task for each classifier was to predict the CEFR level, formulated as a six-way classification problem, of a target word within a given sentence. We report the F1 score for each level, together with Macro-F1 and Micro-F1 scores to measure overall performance.

\begin{table}
\renewcommand{\arraystretch}{1.1}
\centering
\small
\begin{tabular}{crrcrr} \toprule
CEFR & \multicolumn{2}{ c }{EVP Online} & & \multicolumn{2}{ c }{SemCor-CEFR} \\ \cmidrule{2-3} \cmidrule{5-6}
Level & \# types & \# words & & \# types & \# words \\ \midrule
A1 & 577 & 2,932 & & 403 & 31,093\\
A2 & 1,037 & 4,307 & & 680 & 21,065\\
B1 & 1,760 & 7,174 & & 1,206 & 28,707 \\
B2 & 2,368 & 8,754 & & 1,684 & 23,081 \\
C1 & 1,419 & 3,791 & & 849 & 6,701 \\
C2 & 1,692 & 4,604 & & 992 & 5,647 \\ \midrule
Total & 8,853 & 31,562 & & 5,814 & 116,294 \\
\bottomrule\end{tabular}
\caption{Distribution of word types and tokens by CEFR level in the EVP Online and the SemCor-CEFR corpus.}
\label{tab:distribution}
\end{table}

\subsection{Classifiers}
\label{sec:classifiers}

We compared the performance of a baseline model, ME6 Contextual, with several LLM-based approaches, including zero-shot, few-shot, and fine-tuned models.

\paragraph{ME6 Contextual.}
We reimplemented ME6 Contextual as a baseline.
This method uses BERT-based contextual embeddings to train a support vector classifier (SVC) that predicts CEFR levels.
For the hyperparameters of BERT and SVC that were not explicitly specified in the original study~\cite{Aleksandrova:23}, we used their default settings.
Although the model supports MWEs, we excluded them to align with the scope of WordNet, which contains very few MWEs.
We trained three versions of the model: one on 90\% of the EVP examples, one on the SemCor-CEFR corpus, and one on a combination of both. If the model trained on our corpus performs comparably to the model trained on the gold-standard EVP examples, this would support the accuracy of our level annotations.

As noted in Section~\ref{sec:our_wordnet}, a single sense in our corpus may be associated with multiple CEFR levels. Therefore, when training on our data, we created one training example per level.
These multi-level assignments arise when a single WordNet sense aligns with multiple, more fine-grained EVP senses.
Rather than discarding them as noise, we split them into separate training instances—one per level—and treat them as alternative supervision signals.
Because 96.84\% of senses in our WordNet receive a single CEFR label (Table~\ref{tab:cefr-labels-per-sense} in Appendix~\ref{sec:multi_labeled_sense_keys}), we expect the overall impact of this multi-label handling to be limited. Nevertheless, since the EVP evaluation examples provide a single gold level per sentence, this strategy may introduce label ambiguity and reduce performance. Importantly, it should not provide an advantage over models trained solely on EVP data.

\paragraph{Zero-Shot LLM.}
We evaluated the LLM's inherent ability to classify CEFR levels without providing any examples.
Using the prompt and parameter settings described in Appendix~\ref{sec:appendix_a}, we provided the model (GPT-5, checkpoint \texttt{gpt-5-2025-08-07}) with a target word and its context and asked it to output the corresponding CEFR level.

\paragraph{Few-Shot Prompted LLMs.}
We also evaluated the LLM's performance under 6-shot and 18-shot prompting, using the template provided in Appendix~\ref{sec:appendix_a}. 
This prompt provides the model with training examples to serve as clues for classifying a word sense in its context.
In the 6-shot setting, we provided one training example for each of the six CEFR levels, that is, $1 \times 6$ examples. In the 18-shot setting, we used three examples per level, that is, $3 \times 6$ examples.
The target words and usage examples were randomly selected from the 90\% of EVP examples reserved for training.
The LLM and parameter settings were identical to those used in the zero-shot experiments.

\paragraph{Fine-Tuned LLMs.}
We fine-tuned the lightweight and cost-effective GPT-4.1 mini model (checkpoint \texttt{gpt-4.1-mini-2025-04-14}) on three datasets: 90\% of the EVP examples, the SemCor-CEFR corpus, and a combination of both.
As in the evaluation of the ME6 Contextual baseline, the accuracy of our annotations would be supported if the model fine-tuned on our corpus achieved performance comparable to or better than that of the model trained on the gold-standard EVP examples. We also trained a model on the combined corpus to examine potential synergistic effects.
Senses associated with multiple CEFR levels in our corpus were treated as separate training examples for each level.
For fine-tuning, we used a 90\%/10\% split for training and validation. The training data were formatted by filling the zero-shot template in Appendix~\ref{sec:appendix_a} with each target word and sentence, using the corresponding CEFR level as the correct label. The default auto hyperparameters used for fine-tuning are listed in Appendix~\ref{sec:appendix_a}.

\paragraph{Fine-Tuned LLMs + Knowledge Base.}
For words whose CEFR level is unambiguous in the EVP Online, that is, all senses share the same level, performing a full six-level classification is computationally inefficient and increases the risk of errors.
To address this issue, we developed a hybrid approach. We first constructed a knowledge base---a word-level list derived from the EVP Online that includes only words associated with a single CEFR level. For each target word, we checked this list first. If the word was present, we directly assigned its recorded level. Otherwise, we used one of the fine-tuned LLMs for classification.
We applied this method to the LLMs fine-tuned on the EVP examples, the SemCor-CEFR corpus, and the combined corpus to compare differences in classification accuracy.

\begin{table*}[tb]
\small
\renewcommand{\arraystretch}{1.1}
\centering
\begin{tabular}{ccccccccccc} \toprule
\multicolumn{1}{ c }{\multirow{2}{*}{Classifier}} & \multicolumn{1}{ c }{\multirow{2}{*}{Base Model}} & \multicolumn{1}{ c }{Train/Valid.} & \multicolumn{8}{ c }{F1 Scores $\uparrow$} \\ \cmidrule{4-11}
&&Set& \multicolumn{1}{ c }{A1} & \multicolumn{1}{ c }{A2} & \multicolumn{1}{ c }{B1}  & \multicolumn{1}{ c }{B2}  & \multicolumn{1}{ c }{C1}  & \multicolumn{1}{ c }{C2} & \multicolumn{1}{ c }{Macro} & \multicolumn{1}{ c }{Micro}\\ \midrule
\multicolumn{1}{ c }{\multirow{3}{*}{ME6 Cont.}} & \multicolumn{1}{ c }{\multirow{3}{*}{BERT}} & EVP & 0.77 & 0.61 & 0.54 & 0.53 & 0.51 & 0.59 & 0.59 & 0.57 \\
& & SemCor-CEFR & 0.61 & 0.51 & 0.50 & 0.42 & 0.46 & 0.57 & 0.51 & 0.50 \\
& & Mixture & 0.76 & 0.65 & 0.59 & 0.51 & 0.54 & 0.59 & 0.61 & 0.59 \\ \midrule
Zero-Shot & \multicolumn{1}{ c }{\multirow{3}{*}{GPT-5}} & --- & 0.68 & 0.44 & 0.40 & 0.53 & 0.29 & 0.21 & 0.42 & 0.45 \\
6-Shot & & EVP & 0.66 & 0.44 & 0.44 & 0.57 & 0.40 & 0.32 & 0.47 & 0.49 \\
18-Shot & & EVP & 0.67 & 0.45 & 0.43 & 0.56 & 0.38 & 0.40 & 0.48 & 0.49 \\ \midrule
\multicolumn{1}{ c }{\multirow{3}{*}{FT}} & \multicolumn{1}{ c }{\multirow{3}{*}{GPT-4.1 mini}} & EVP & 0.79 & 0.68 & 0.64 & 0.69 & 0.43 & 0.68 & 0.65 & 0.66 \\
& & SemCor-CEFR & 0.72 & 0.67 & 0.68 & 0.71 & 0.44 & 0.66 & 0.67 & 0.67 \\
& & Mixture & \underline{0.81} & 0.76 & 0.73 & 0.75 & 0.61 & 0.73 & 0.73 & 0.73 \\ \midrule
\multicolumn{1}{ c }{\multirow{3}{*}{FT+KB}} & \multicolumn{1}{ c }{\multirow{3}{*}{GPT-4.1 mini}} & EVP & \textbf{0.83} & \underline{0.77} & 0.74 & 0.79 & 0.74 & \textbf{0.81} & \underline{0.78} & \underline{0.78} \\
& & SemCor-CEFR & 0.75 & 0.72 & \underline{0.76} & \underline{0.81} & \underline{0.75} & \underline{0.77} & 0.76 & 0.76 \\
& & Mixture & \textbf{0.83} & \textbf{0.81} & \textbf{0.78} & \textbf{0.83} & \textbf{0.78} & \textbf{0.81} & \textbf{0.81} & \textbf{0.81} \\ \bottomrule
\end{tabular}
\caption{F1 scores for each classifier. \textbf{Bold} and \underline{underlined} values indicate the highest and second-highest scores, respectively.}
\label{tab:F1_scores}
\end{table*}
\subsection{Results}

Table~\ref{tab:F1_scores} reports the F1 scores for each classifier. In the table, FT denotes the fine-tuned LLMs, and FT+KB refers to the fine-tuned LLMs combined with the knowledge-based approach.
The training datasets used are EVP (90\% of the EVP examples), SemCor-CEFR (our annotated SemCor corpus), and Mixture (a combination of both).
Because the class distribution in our data is imbalanced, we use the Macro-F1 score as the primary metric for overall evaluation, as it assigns equal weight to each class and mitigates the effects of frequency imbalance.

The ME6 Contextual classifier achieved a Macro-F1 score of at least 0.5 across all training sets. 
However, its performance on the SemCor-CEFR corpus was 0.08 points lower than that on the EVP data.
We attribute this gap to the model's vector construction method, which averages the vectors for all instances of a given word and CEFR level to produce a single vector for each word-level pair.
As shown in Table~\ref{tab:distribution}, our SemCor-CEFR corpus has fewer unique word types than the EVP data.
Consequently, despite having a higher total word frequency, it yields fewer training vectors, which likely contributed to the performance drop. 
Consistent with this interpretation, the classifier trained on the Mixture dataset, which included the largest number of training examples, achieved the best performance among the ME6 Contextual models.

The zero-shot LLM achieved a Macro-F1 score of 0.42, the lowest among all methods and well below that of the ME6 Contextual baseline. Its F1 scores for the C1 and C2 levels were particularly low, below 0.3, indicating that the LLM's internal knowledge alone is insufficient for classifying advanced-level senses.
Providing in-context examples through few-shot prompting increased the Macro-F1 score to 0.47 in the 6-shot setting and 0.48 in the 18-shot setting. This improvement, consistent with prior findings~\cite{Enomoto:24,Smadu:24}, resulted from supplementing the model's knowledge of C1 and C2 senses.
 Nevertheless, the performance of the few-shot models remained substantially lower than that of ME6 Contextual.

Fine-tuning proved to be a highly effective approach for developing LLM classifiers, improving the Macro-F1 score by at least 0.17 points compared with the few-shot methods.
Notably, the FT model trained on the SemCor-CEFR corpus performed comparably to the model trained on the gold-standard EVP data, despite being optimized on a dataset entirely different from the test set. Moreover, an analysis of its errors (Figure~\ref{fig:ft_semcor} in Appendix~\ref{sec:appendix_b}) shows that misclassifications, particularly for C1-level senses, were often assigned to adjacent proficiency levels, which would likely minimize confusion for learners. 
This strong performance is likely due to the model being fine-tuned on the rich and varied usage examples in the SemCor-CEFR corpus.
The model trained on the Mixture dataset achieved a Macro-F1 score of 0.73. 
These results provide indirect evidence supporting the quality of the CEFR annotations in our WordNet and effectiveness of combining the EVP and SemCor-CEFR corpora.

The hybrid FT+KB approach, which combines fine-tuned LLMs with a knowledge base, achieved the best overall performance. This method improved the Macro-F1 score by 0.08 to 0.13 points compared with the FT models alone, with a consistent trend across training sets. 
The classifier trained on the Mixture dataset achieved the highest F1 scores at all levels, exceeding 0.8 for every level except B1 and C1. 
This pattern suggests that a substantial portion of the test set consists of words with unambiguous CEFR levels. 
In such cases, the knowledge base can assign the correct level without relying on LLM inference, thereby improving both accuracy and computational efficiency.

\section{Discussion}

\subsection{Correlation Analysis Using the CompLex 2.0 Dataset}
\label{sec:correlation_analysis}

Although the FT and FT+KB classifiers trained on EVP examples demonstrated strong performance, these results may be inflated because both the fine-tuning and test sets were drawn from the same source.
For real-world applications, a CEFR-level classifier must perform well across diverse domains, not only on dictionary and learner examples. However, gold-standard sense-level CEFR annotations for heterogeneous corpora are scarce.
To examine generalizability, we used an indirect proxy by analyzing the correlation between the predicted CEFR levels and lexical complexity scores in the CompLex 2.0 dataset. This dataset, developed for the LCP task, spans three genres---Europarl, the Bible, and biomedical texts---and contains target words rated by multiple annotators on a continuous complexity scale from 0 to 1.
We applied our classifiers to predict CEFR levels, mapped to integers 1 to 6, for 7,662 target words in the CompLex 2.0 training set. We then computed the Spearman rank correlation coefficient between the predicted levels and the dataset's complexity scores. We did not expect a high correlation because complexity scores are continuous, whereas CEFR levels are discrete.

\begin{table}[tb]
\renewcommand{\arraystretch}{1.1}
\centering
\begin{tabular}{ccc} \toprule
\multicolumn{1}{ c }{Classifier} & \multicolumn{1}{ c }{Train/Valid. Set} & \multicolumn{1}{ c }{Spearman $\uparrow$} \\ \midrule
\multicolumn{1}{ c }{\multirow{3}{*}{ME6 Cont.}} & EVP & 0.333 \\
& SemCor-CEFR & 0.377 \\
& Mixture & 0.362 \\ \midrule
Zero-Shot & --- & 0.396 \\
6-Shot & EVP & 0.494 \\
18-Shot & EVP & 0.490 \\ \midrule
\multicolumn{1}{ c }{\multirow{3}{*}{FT}} & EVP & 0.288 \\
& SemCor-CEFR & \textbf{0.541} \\
& Mixture & 0.529 \\ \midrule
\multicolumn{1}{ c }{\multirow{3}{*}{FT+KB}} & EVP & 0.366 \\
& SemCor-CEFR & \underline{0.539} \\
& Mixture & 0.528 \\ \bottomrule
\end{tabular}
\caption{Spearman rank correlation coefficients between predicted CEFR levels and CompLex 2.0 complexity scores.
\textbf{Bold} and \underline{underlined} values indicate the highest and second-highest scores, respectively.
}
\label{tab:spearman}
\end{table}
A notable finding in Table~\ref{tab:spearman} is that classifiers trained on the EVP examples, despite achieving high accuracy on the EVP test set, exhibited very low correlation with the CompLex 2.0 scores.
This suggests that models fine-tuned solely on EVP data may have learned superficial, dataset-specific cues, with limited transfer to other genres.
By contrast, classifiers trained on the SemCor-CEFR corpus achieved correlation coefficients above 0.5, indicating a moderate relationship between their predictions and lexical complexity. We attribute this improvement to the broad genre coverage of the SemCor corpus together with the high quality of our CEFR-level annotations. 
These findings indicate that classifiers fine-tuned on our corpus are better suited for application to educational materials drawn from diverse sources.

\subsection{Implications for L2 Learners}

Our findings have important implications for L2 learners, who often struggle with the fine-grained sense distinctions in WordNet. 
By annotating WordNet senses with CEFR levels and integrating them into resources such as the SemCor-CEFR corpus, our approach aligns lexical information more closely with learner proficiency and pedagogical needs.
Although the practical benefits of this approach require empirical validation through classroom-based or longitudinal studies, it offers two main advantages.
First, it allows learners to focus on senses that match their proficiency level, reducing the cognitive load associated with more advanced or nuanced meanings. 
Second, our high-performing classifier (Macro-F1 of 0.81) can be integrated into educational tools to quickly identify complex lexical items in a text, enabling immediate scaffolding. 
The model's strong performance on levels A1 through B2 is particularly beneficial for beginner and intermediate learners who are building foundational vocabulary. These advances may enable educators and independent learners to adopt more adaptive and efficient strategies for vocabulary instruction. However, further research is needed to determine whether these benefits persist across diverse learning environments and over extended periods.

\section{Conclusions}

In this study, we introduced an LLM-based method for annotating WordNet senses with CEFR levels and used it to construct a CEFR-annotated WordNet. This resource provides 10,995 proficiency annotations for 10,644 senses across 5,645 lemmas.
Using the annotated WordNet, we also created the SemCor-CEFR corpus, a large-scale resource containing more than 110,000 sense-level CEFR annotations. To validate our approach, we trained contextual lexical CEFR-level classifiers on this corpus and found that they performed comparably to models trained on gold-standard data.
Furthermore, by combining our corpus with gold-standard levels, we developed a practical classifier that achieves a Macro-F1 score of 0.81, providing indirect evidence of the utility and consistency of our CEFR annotations. Our analysis also showed that the predictions generated by our classifiers correlate with the lexical complexity scores in the CompLex 2.0 dataset, suggesting moderate alignment with lexical complexity across diverse text genres.

This work is part of the ``Learner's WordNet Project,'' which seeks to integrate NLP methods---particularly WordNet's rich semantic network---with educational technology to support more efficient and effective L2 learning. 
Future work will focus on expanding CEFR-level coverage in our WordNet, evaluating its pedagogical effectiveness in real-world learning contexts and developing related applications. To support this effort, we plan to build a classifier capable of accurately assigning CEFR levels to previously unannotated word senses.

\section{Acknowledgements}

This work was supported in part by JSPS KAKENHI Grant Numbers JP22K02825, JP22K18006, JP25K21351, and JP24K03052.
This publication/presentation/research report makes use of the English Vocabulary Profile. This resource is based on extensive research using the Cambridge Learner Corpus and forms part of the English Profile program, which aims to provide evidence about language use to support the development of improved language teaching materials.
See \url{https://englishprofile.org/} for more information.

\section{Ethical Considerations}

We accessed the EVP Online strictly in accordance with the EVP website's Terms of Use\footnote{\url{https://englishprofile.org/?menu=evp-terms-of-use}} and Cambridge University Press's text and data mining (TDM) terms\footnote{\url{https://www.cambridge.org/us/legal/copyright}}.
Any EVP content temporarily cached to local storage for this project was deleted upon the project's completion.
All released artifacts are built on WordNet and SemCor, whose licenses permit copying, modification, and redistribution, and include only derived CEFR labels mapped to WordNet sense keys.
No verbatim EVP content, including entries, examples, glosses, or metadata, is included in the released resources.
The EVP-derived word-level lookup list used in the FT+KB analysis (Section~\ref{sec:classifiers}) was used solely for internal evaluation and is not redistributed.

For verification, we employed proprietary LLMs from OpenAI and enabled the opt-out setting to ensure that submitted data were not used for model training.
For downstream applications involving personal or sensitive data, we recommend deploying open-source LLMs in a local environment to reduce the risk of unintended disclosure.
The resources introduced in this work are compatible with locally hosted LLMs. 
The CEFR levels provided by our resources are model-based estimates and should not be used as the sole basis for high-stakes educational decisions, such as promotion, pass/fail determinations, or selective admissions.

\section{Limitations}

A key limitation of this work is the limited coverage of CEFR-level annotations in WordNet.
Despite the automated pipeline, only 10,644 senses were annotated, representing approximately 5\% of the full inventory. This limited coverage stems from the restricted availability of sense-level CEFR labels in the EVP Online, which serves as the primary source for alignment and constrains further expansion.
Beyond the EVP Online, several CEFR-related lexical resources are available, such as CEFRLex and SeCoDa, as well as lexical complexity datasets including CompLex 2.0.
However, these resources differ in granularity and label space. Proficiency information is often provided at the word level rather than the sense level, sense inventories are not aligned with WordNet, and labels are not always expressed as CEFR levels. As a result, they cannot be directly integrated into our sense-level annotations without additional mapping. Developing principled methods to use them as auxiliary supervision remains future work.
To mitigate the limited coverage, we developed lexical CEFR-level classifiers trained on the large, sense-annotated corpus, achieving a maximum Macro-F1 score of 0.81. Although these classifiers can predict levels for previously unseen senses based on contextual usage, they are currently less accurate than the primary gloss-based transfer method. Improving their accuracy, robustness, and generalization is therefore essential for reliable large-scale deployment. In parallel, it is important to assess how annotation errors affect L2 learners and establish acceptable error thresholds for educational applications.
Another limitation is that the evaluation focused exclusively on single words, whereas related work on LCP, including models such as ME6 Contextual, also considers multiword expressions.
Preliminary analysis suggests that most MWEs correspond to a single CEFR level, indicating that the knowledge-based component could classify them with high accuracy. However, challenges remain in identifying implicit MWEs in running text, such as in CALL systems analyzing textbooks, and in handling expressions not covered by the EVP. Addressing these issues will require more advanced and context-sensitive classification methods.

\nocite{*}
\section{Bibliographical References}\label{sec:reference}

\bibliographystyle{lrec2026-natbib}
\bibliography{refs}

\section{Language Resource References}
\label{lr:ref}
\bibliographystylelanguageresource{lrec2026-natbib}
\bibliographylanguageresource{languageresource}
\clearpage

\section{Appendices}

\begin{figure}[tb]
\centering
{\setlength{\fboxsep}{2mm}\fbox{\parbox{7.2cm}{
Please assess whether the two meanings of the English word \{word\} are the same from a linguistic perspective.\\
1: \{one gloss $g$ of \{word\} in the EVP Online\}\\
2: \{one gloss $g'$ of \{word\} in WordNet\}\\
\\
Please select one option from the following and answer using only the corresponding number.\\
1. Exactly the same meaning\\
2. Almost the same meaning\\
3. Somewhat similar meaning\\
4. Neither similar nor different meaning\\
5. Somewhat different meaning\\
6. Mostly different meaning\\
7. Completely different meaning
}}}
\caption{Prompt template used to measure semantic similarity between an EVP gloss ($g$) and a WordNet gloss ($g'$).}
\label{fig:similarity_measurement}
\end{figure}
\begin{figure}[tb]
\centering
{\setlength{\fboxsep}{2mm}\fbox{\parbox{7.2cm}{
The CEFR is a six-level scale, with each level corresponding to a specific English proficiency level. The levels are as follows:\\
\\
A1: Beginner\\
A2: Elementary\\
B1: Intermediate\\
B2: Upper Intermediate\\
C1: Advanced\\
C2: Proficiency\\
\\
According to the CEFR scale, which proficiency level is required to understand the sense of \{word\} in the following text: \{sentence\}\\
\\
Please do not provide explanations.
}}}
\caption{Prompt template used for zero-shot CEFR-level classification.}
\label{fig:zero-shot_prompt}
\end{figure}
\begin{figure*}[tb]
\centering
{\setlength{\fboxsep}{2mm}\fbox{\parbox{15.5cm}{
The CEFR is a six-level scale, with each level corresponding to a specific English proficiency level. The levels are as follows:\\
\\
A1: Beginner\\
A2: Elementary\\
B1: Intermediate\\
B2: Upper Intermediate\\
C1: Advanced\\
C2: Proficiency\\
\\
According to the CEFR scale, the proficiency levels required to understand the senses of the words in the following texts are:\\
\\
Word: \{train\_word$_1$\}, Text: \{train\_sentence$_1$\} -> CEFR: \{The gold-standard level $\ell_1$\}\\
Word: \{train\_word$_2$\}, Text: \{train\_sentence$_2$\} -> CEFR: \{The gold-standard level $\ell_2$\}\\
\\
(...more training examples...)\\
\\
Word: \{test\_word\}, Text: \{test\_sentence\} -> CEFR:\\
\\
Please respond with only the level.
}}}
\caption{Prompt template used for few-shot CEFR-level classification.}
\label{fig:few-shot_prompt}
\end{figure*}
\subsection{Parameters and Prompts}
\label{sec:appendix_a}

To ensure reproducibility, we report the exact checkpoint of the OpenAI model used in this study.
Different checkpoints were used for gloss similarity measurements (Section~\ref{sec:our_wordnet}) and classifier experiments (Section~\ref{sec:experiment}), as these components were conducted at different times. For each task, we used the most recent checkpoint available at the time of execution.

\paragraph{Semantic Similarity Measurement.}
To measure the semantic similarity between $g_i$ and $g'_j$, we used GPT-4o (checkpoint \texttt{gpt-4o-2024-08-06}) with the prompt shown in Figure~\ref{fig:similarity_measurement}.
The system message was set at ``You are a professional linguist,'' and the temperature was set at 0 to ensure deterministic outputs.

\paragraph{Zero-Shot and Few-Shot Classifiers.}
We used GPT-5 (checkpoint \texttt{gpt-5-2025-08-07}) as the base model for our classifiers. The system message was set at ``You are an expert rater for the Common European Framework of Reference for Languages (CEFR).'' and the parameter reasoning\_effort was set at ``high.''
Figures~\ref{fig:zero-shot_prompt} and~\ref{fig:few-shot_prompt} present the prompt templates used for the zero-shot and few-shot LLM classifiers, respectively.
In a preliminary experiment, we provided the LLMs with full CEFR level descriptions based on the official can-do descriptors.
However, we observed no significant difference in classification performance compared to using the simplified descriptions shown in the figures.
For efficiency, we therefore used the prompts with simplified descriptions in our experiments.

\paragraph{Fine-Tuned LLMs.}
As described in Section~\ref{sec:classifiers}, we constructed the training data using the zero-shot template shown in Figure~\ref{fig:zero-shot_prompt}.
The hyperparameters used for fine-tuning GPT-4.1 mini (checkpoint \texttt{gpt-4.1-mini-2025-04-14}) are detailed in Table~\ref{tab:parameters}.

\begin{table}[tb]
\renewcommand{\arraystretch}{1.1}
\centering
\begin{tabular}{clr} \toprule
\multicolumn{1}{ c }{Train/Valid. Set} & \multicolumn{1}{ c }{Parameter} & \multicolumn{1}{ c }{Value} \\ \midrule
\multirow{5}{*}{EVP} & Method & Supervised \\
& Seed & 1900973879\\
& Batch size & 17\\
& LR multiplier & 2\\
& Epochs & 1\\ \midrule
\multirow{5}{*}{SemCor-CEFR} & Method & Supervised \\
& Seed & 105188566\\
& Batch size & 69\\
& LR multiplier & 2\\
& Epochs & 1\\ \midrule
\multirow{5}{*}{Mixture} & Method & Supervised \\
& Seed & 112279849\\
& Batch size & 86\\
& LR multiplier & 2\\
& Epochs & 1\\ \bottomrule
\end{tabular}
\caption{Hyperparameters for the fine-tuned LLMs.}
\label{tab:parameters}
\end{table}

\begin{table}[tb]
\renewcommand{\arraystretch}{1.1}
\centering
\small
\begin{tabular}{@{}rrr@{}}
\toprule
\# Distinct CEFR Labels & \# Sense Keys & Share (\%) \\
\midrule
1 & 10,308 & 96.84 \\
2 & 321    & 3.02  \\
3 & 15     & 0.14  \\
\bottomrule
\end{tabular}
\caption{Distribution of the number of distinct CEFR labels per WordNet sense key.}
\label{tab:cefr-labels-per-sense}
\end{table}

\begin{table}[tb]
\renewcommand{\arraystretch}{1.1}
\centering
\begin{tabular}{@{}lrrrrrr@{}}
\toprule
\multirow{2}{*}{Row Level} & \multicolumn{6}{c}{Column Level} \\
\cmidrule{2-7}
 & A1 & A2 & B1 & B2 & C1 & C2 \\
\midrule
A1 & ---  & 33 & 31 & 13 & 8  & 2  \\
A2 & 33 & ---  & 43 & 24 & 4  & 6  \\
B1 & 31 & 43 & ---  & 86 & 12 & 13 \\
B2 & 13 & 24 & 86 & ---  & 24 & 40 \\
C1 & 8  & 4  & 12 & 24 & ---  & 27 \\
C2 & 2  & 6  & 13 & 40 & 27 & ---  \\
\bottomrule
\end{tabular}
\caption{Pairwise co-occurrence counts of CEFR levels within multi-labeled sense keys. Each cell $(x,y)$ reports the number of sense keys whose label set contains both levels $x$ and $y$; three-level cases contribute to multiple pairs.}
\label{tab:cefr-cooccurrence}
\end{table}

\begin{table*}[tb]
\renewcommand{\arraystretch}{1.1}
\centering
\small
\setlength{\tabcolsep}{4pt}
\begin{tabularx}{\textwidth}{@{}l l l c Y@{}}
\toprule
\multicolumn{1}{c}{Sense Key} & \multicolumn{1}{c}{Lemma} & \multicolumn{1}{c}{PoS} & \multicolumn{1}{c}{CEFR (3 Labels)} & \multicolumn{1}{c}{WordNet Gloss (Definition)} \\
\midrule
\texttt{bad\%3:00:00::} & bad & Adjective & A1/A2/C1 & having undesirable or negative qualities \\
\texttt{block\%2:35:02::} & block & Verb & B2/C1/C2 & obstruct \\
\texttt{close\%2:41:00::} & close & Verb & A2/B2/C2 & cease to operate or cause to cease operating \\
\texttt{dance\%1:04:00::} & dance & Noun & A1/A2/B1 & taking a series of rhythmical steps (and movements) in time to music \\
\texttt{find\%2:39:02::} & find & Verb & A1/A2/B1 & discover or determine the existence, presence, or fact of \\
\texttt{give\%2:32:02::} & give & Verb & A1/A2/B1 & bestow \\
\texttt{give\%2:36:00::} & give & Verb & A1/A2/B1 & give or supply \\
\texttt{hard\%3:00:06::} & hard & Adjective & A1/B1/C1 & not easy \\
\texttt{miss\%2:32:00::} & miss & Verb & A2/B1/B2 & fail to experience \\
\texttt{safe\%3:00:01::} & safe & Adjective & A1/A2/B1 & free from danger or the risk of harm \\
\texttt{schedule\%1:10:00::} & schedule & Noun & A2/B1/B2 & an ordered list of times at which things are planned to occur \\
\texttt{shake\%2:29:00::} & shake & Verb & B1/B2/C2 & move with or as if with a tremor \\
\texttt{start\%2:36:01::} & start & Verb & A1/B1/B2 & get off the ground \\
\texttt{start\%2:38:00::} & start & Verb & A1/B1/B2 & begin or set in motion \\
\texttt{start\%2:38:01::} & start & Verb & A1/B1/B2 & get going or set in motion \\
\bottomrule
\end{tabularx}
\caption{WordNet sense keys annotated with three distinct CEFR levels. For each sense key, we report the lemma, PoS, and the WordNet gloss.}
\label{tab:three-level-sensekeys}
\end{table*}
\subsection{Multi-Labeled Sense Keys}
\label{sec:multi_labeled_sense_keys}

In WordNet, a sense key is the canonical string identifier of an individual sense, encoding the lemma, PoS, and a sense index.
Because our CEFR annotation transfers levels from EVP senses to WordNet senses via gloss alignment, a single WordNet sense key may occasionally receive multiple CEFR labels. This occurs when a WordNet gloss is sufficiently similar, with a similarity score of $\leq 2$, to more than one EVP gloss and those EVP glosses carry different CEFR levels.
Such multi-label assignments primarily reflect differences in gloss granularity between the two resources.

\paragraph{How frequent are multi-labeled sense keys?}
Table~\ref{tab:cefr-labels-per-sense} summarizes the number of distinct CEFR labels assigned to each WordNet sense key. The vast majority of sense keys are unambiguous: 10,308 sense keys (96.84\%) receive exactly one label.
Multi-labeled cases are rare, with 321 sense keys (3.02\%) assigned two labels and only 15 sense keys (0.14\%) assigned three labels.
 Overall, multi-label assignments affect approximately 3.16\% of all annotated sense keys. This indicates that the conservative alignment threshold produces predominantly single-level annotations while retaining coverage for borderline cases.

\paragraph{Which level combinations co-occur?}
To characterize the nature of multi-labeling, Table~\ref{tab:cefr-cooccurrence} reports pairwise co-occurrence counts of CEFR levels within the multi-labeled sense keys.
Two-label cases contribute one pair, and three-label cases contribute three pairs, yielding 366 total co-occurrence pairs in Table~\ref{tab:cefr-cooccurrence}.
A clear pattern emerges. The most frequent co-occurrence is B1--B2 (86), followed by A2--B1 (43), B2--C2 (40), and A1--A2 (33).
Overall, adjacent-level pairs (A1--A2, A2--B1, B1--B2, B2--C1, C1--C2) dominate. This pattern suggests that multi-labeling typically occurs near proficiency boundaries rather than arising from arbitrary mismatches.
Less frequent and more distant pairs, such as A1--C1 and A2--C2, may reflect particularly broad WordNet glosses or EVP senses whose pedagogical sequencing differs substantially across sub-senses.

\paragraph{What do three-label cases look like?}
Table~\ref{tab:three-level-sensekeys} lists the 15 sense keys assigned three distinct CEFR labels.
These cases are dominated by highly frequent and semantically broad lemmas (e.g., \textit{bad}, \textit{close}, \textit{find}, \textit{give}, \textit{start}), which are associated with short and general WordNet glosses.
Such senses can plausibly align with EVP sub-senses introduced at different stages, for example early concrete uses versus later, more abstract or specialized uses.
Importantly, the extremely small number of three-label cases suggests that wide label dispersion is exceptional; most multi-labeled sense keys involve only two nearby levels.

\begin{figure*}[tb]
    \begin{minipage}[b]{0.33 \linewidth}
    \centering
    \includegraphics[keepaspectratio, scale=0.22]{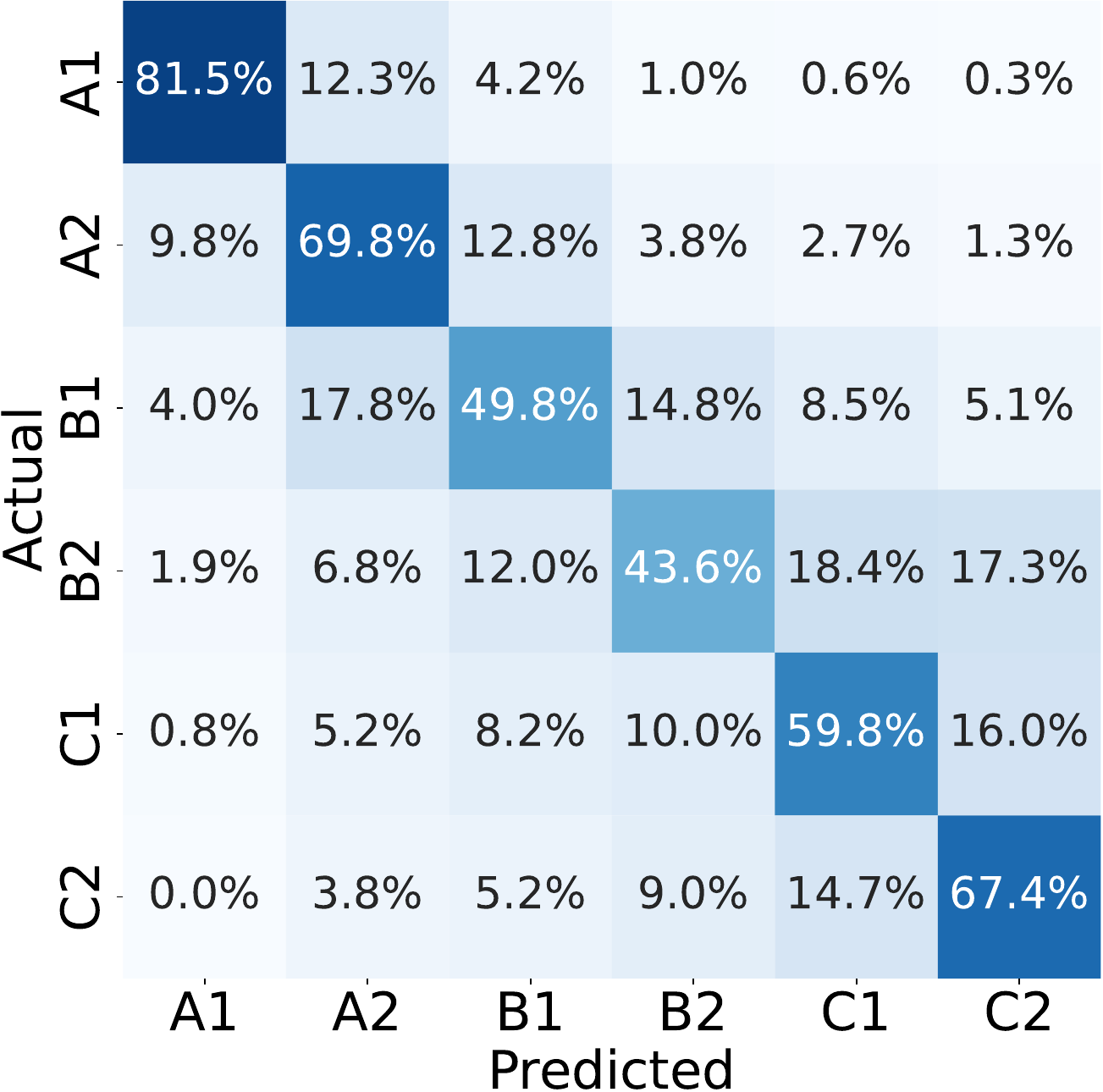}
    \subcaption{ME6 Cont. using EVP.}\label{fig:me6_evp}
    \end{minipage}
    \begin{minipage}[b]{0.33 \linewidth}
    \centering
    \includegraphics[keepaspectratio, scale=0.22]{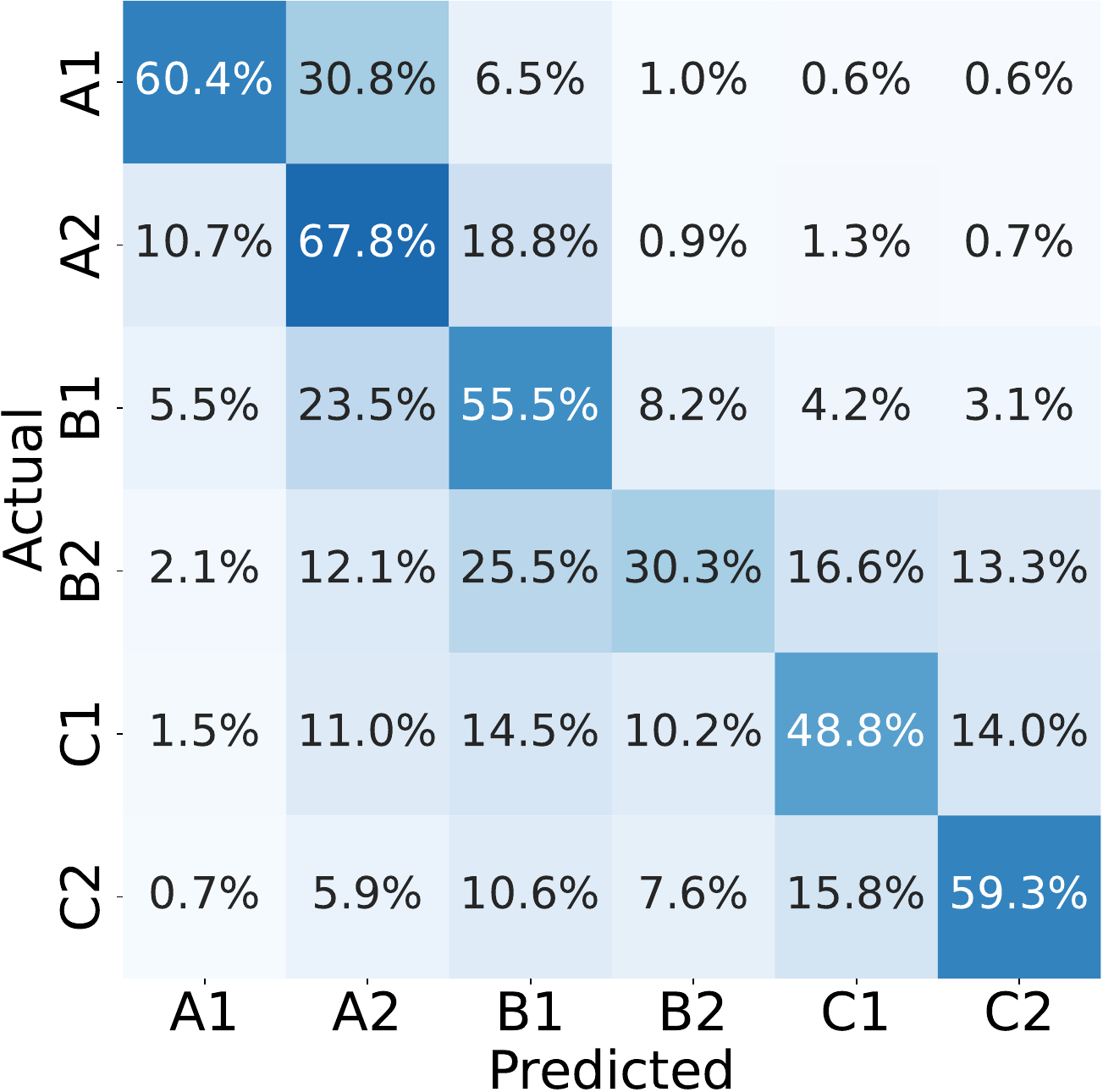}
    \subcaption{ME6 Cont. using SemCor-CEFR.}\label{fig:me6_semcor}
    \end{minipage}
    \begin{minipage}[b]{0.33 \linewidth}
    \centering
    \includegraphics[keepaspectratio, scale=0.22]{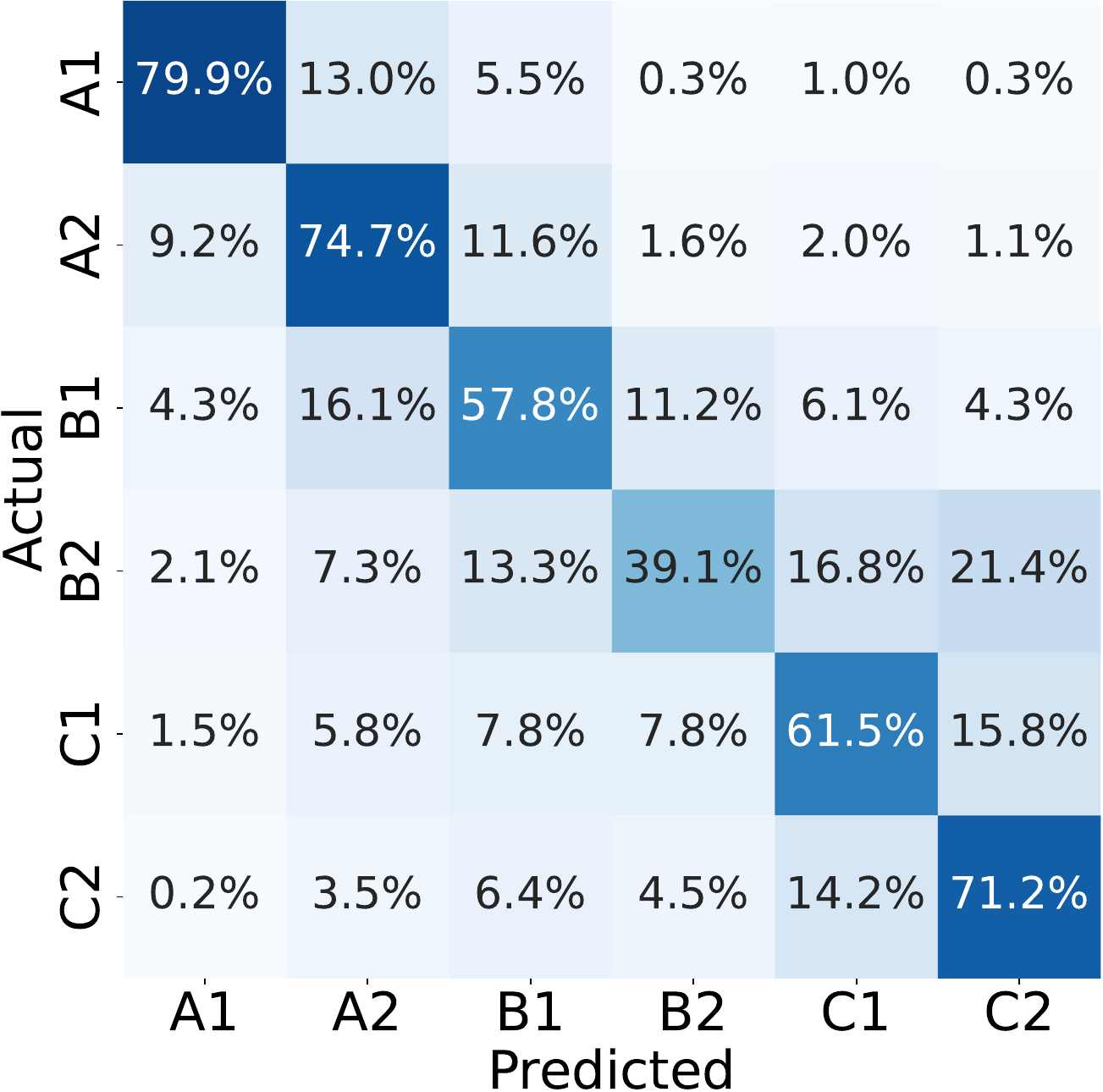}
    \subcaption{ME6 Cont. using Mixture.}\label{fig:me6_evp+semcor}
    \end{minipage}\\\\
    \begin{minipage}[b]{0.33 \linewidth}
    \centering
    \includegraphics[keepaspectratio, scale=0.22]{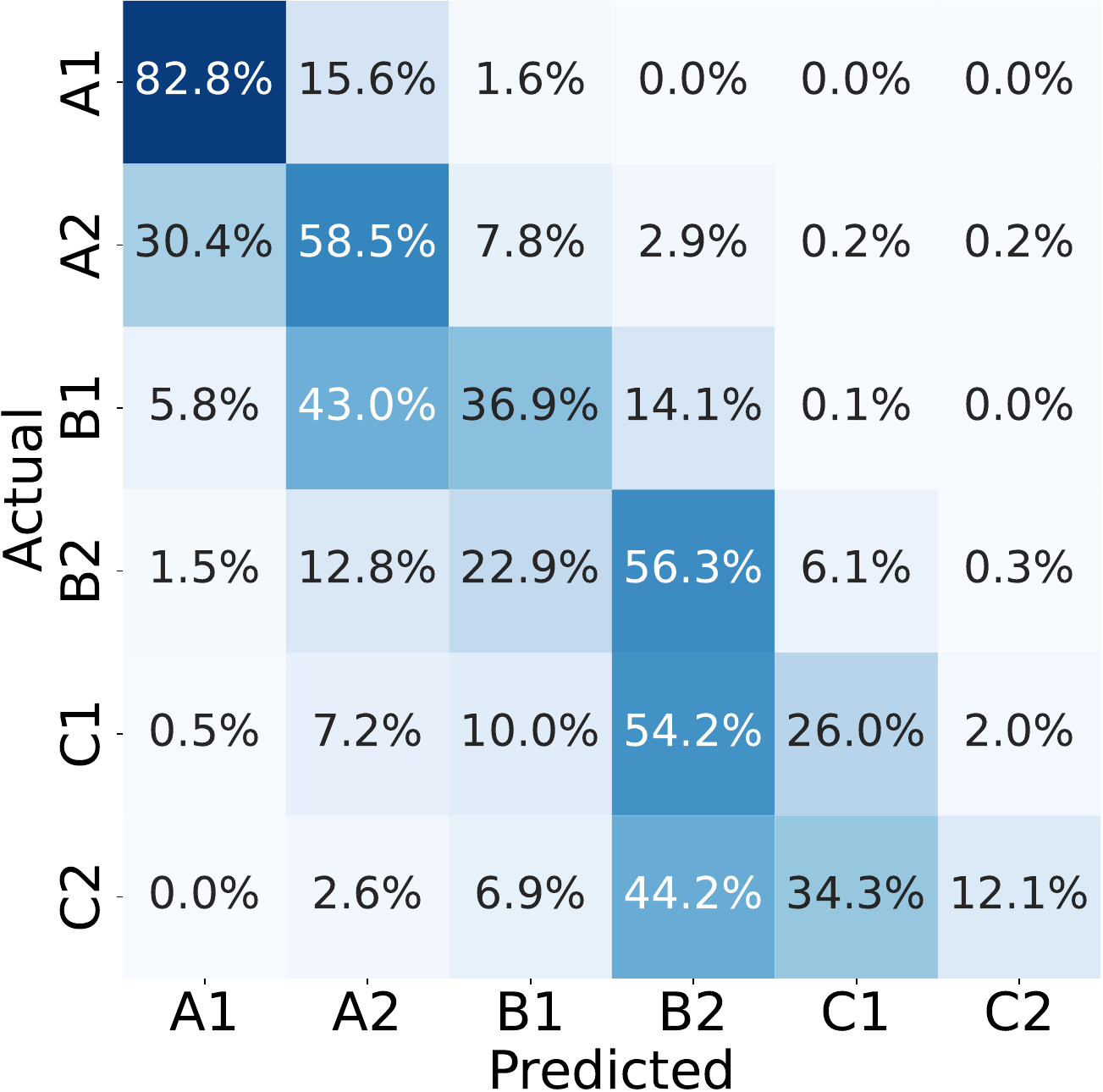}
    \subcaption{Zero-Shot.}\label{fig:zero-shot}
    \end{minipage}
    \begin{minipage}[b]{0.33 \linewidth}
    \centering
    \includegraphics[keepaspectratio, scale=0.22]{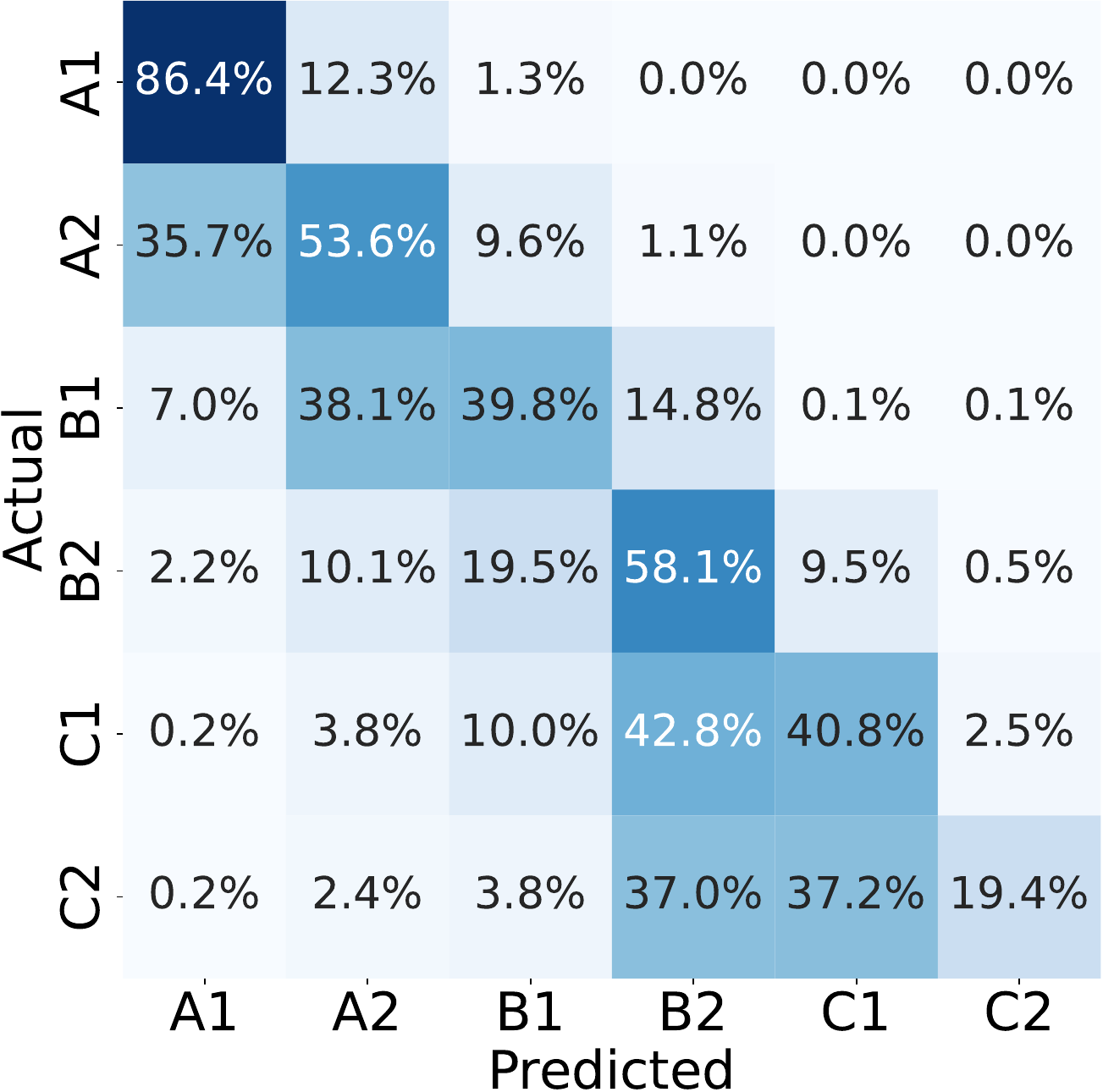}
    \subcaption{6-Shot.}\label{fig:6-shot}
    \end{minipage}
    \begin{minipage}[b]{0.33 \linewidth}
    \centering
    \includegraphics[keepaspectratio, scale=0.22]{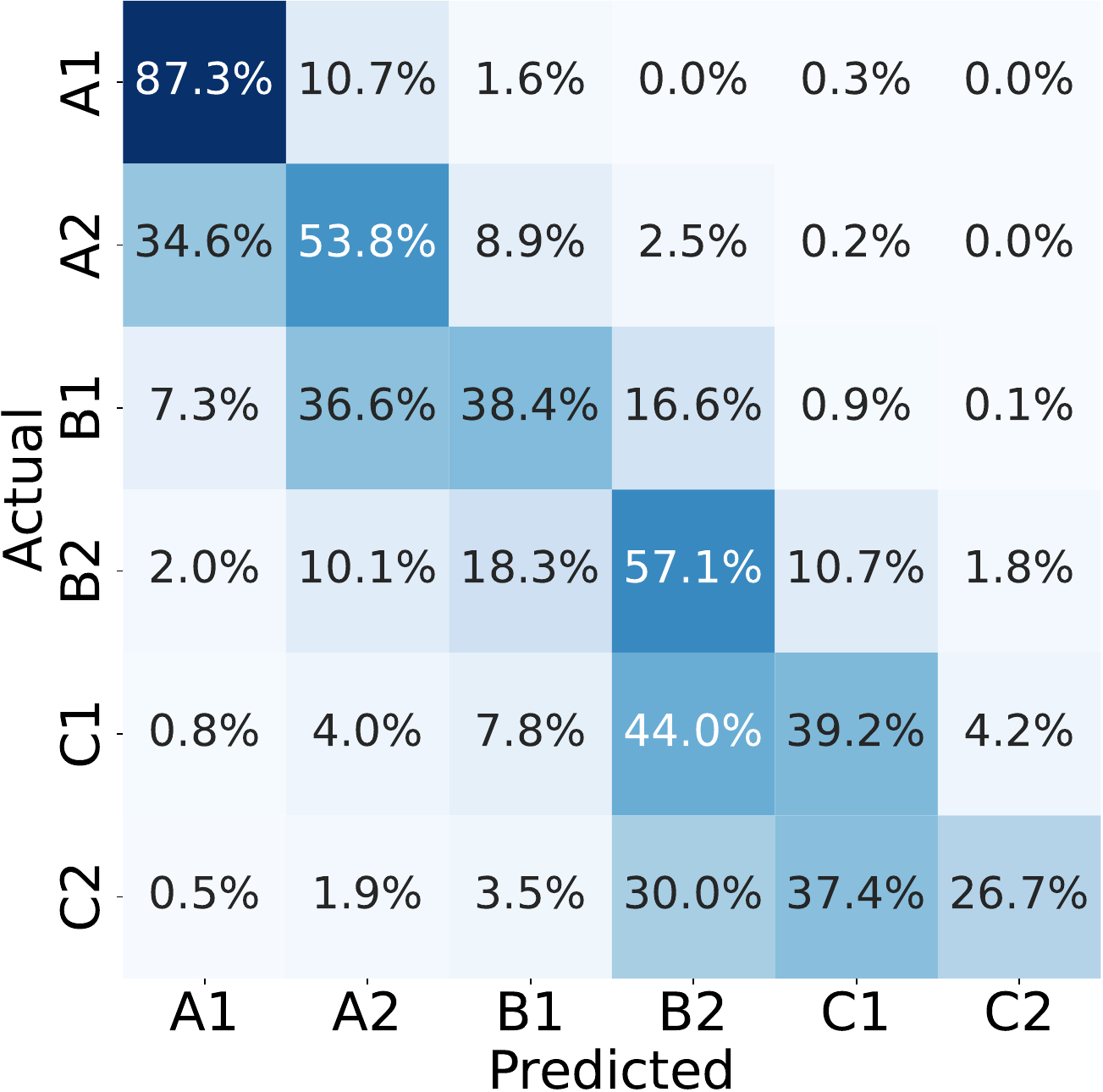}
    \subcaption{18-Shot.}\label{fig:18-shot}
    \end{minipage}\\\\
    \begin{minipage}[b]{0.33 \linewidth}
    \centering
    \includegraphics[keepaspectratio, scale=0.22]{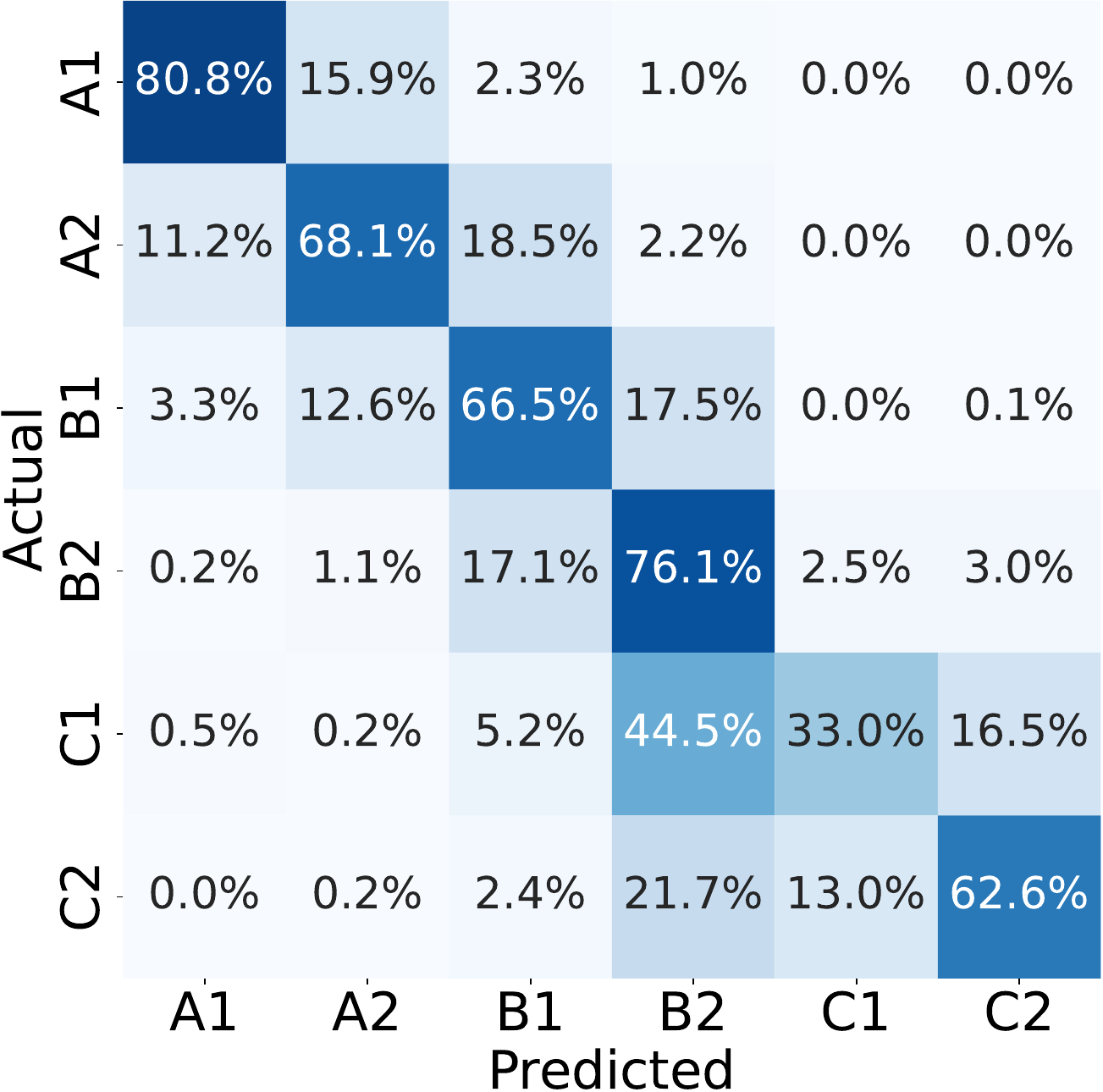}
    \subcaption{FT using EVP.}\label{fig:ft_evp}
    \end{minipage}
    \begin{minipage}[b]{0.33 \linewidth}
    \centering
    \includegraphics[keepaspectratio, scale=0.22]{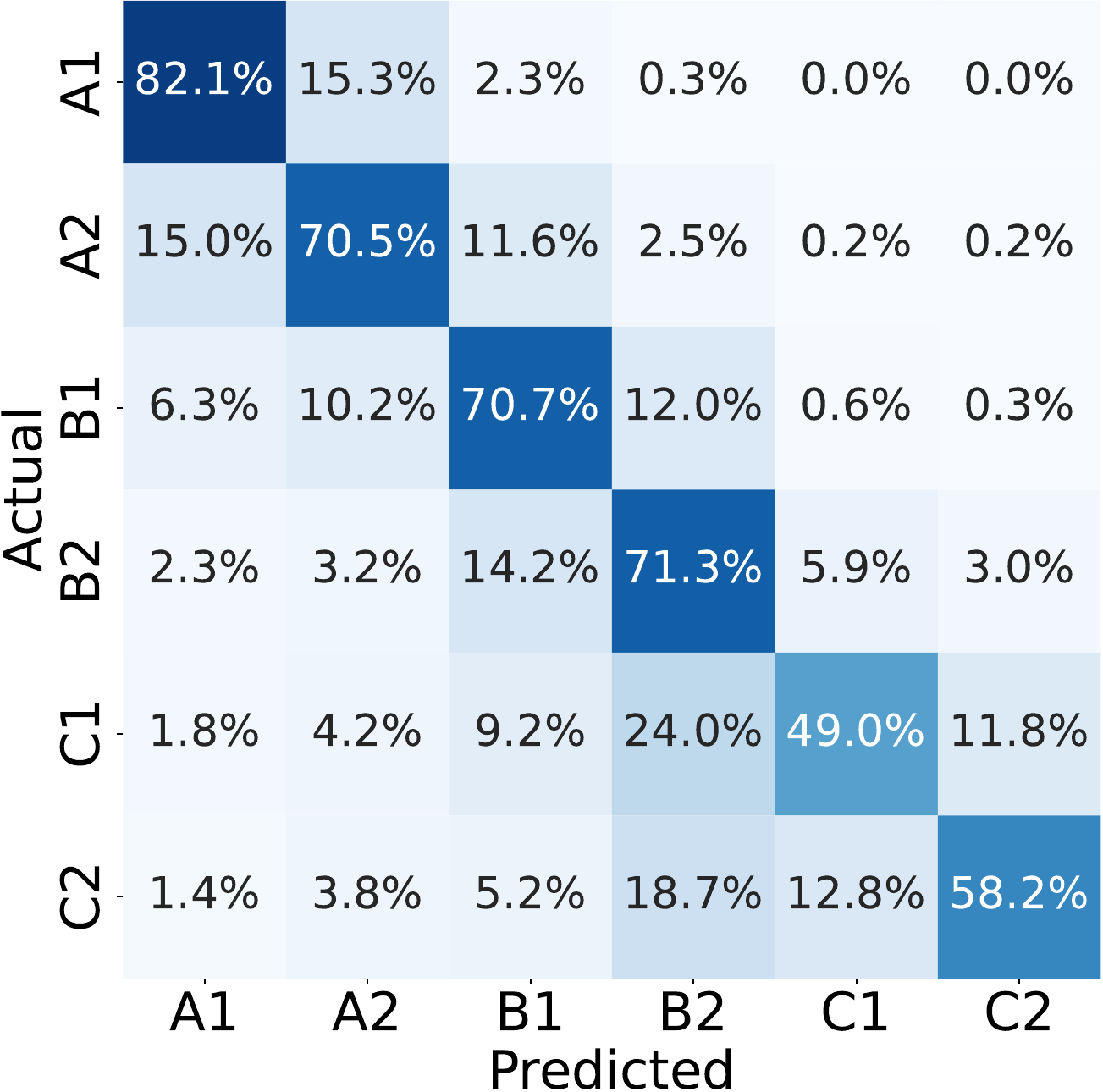}
    \subcaption{FT using SemCor-CEFR.}\label{fig:ft_semcor}
    \end{minipage}
    \begin{minipage}[b]{0.33 \linewidth}
    \centering
    \includegraphics[keepaspectratio, scale=0.22]{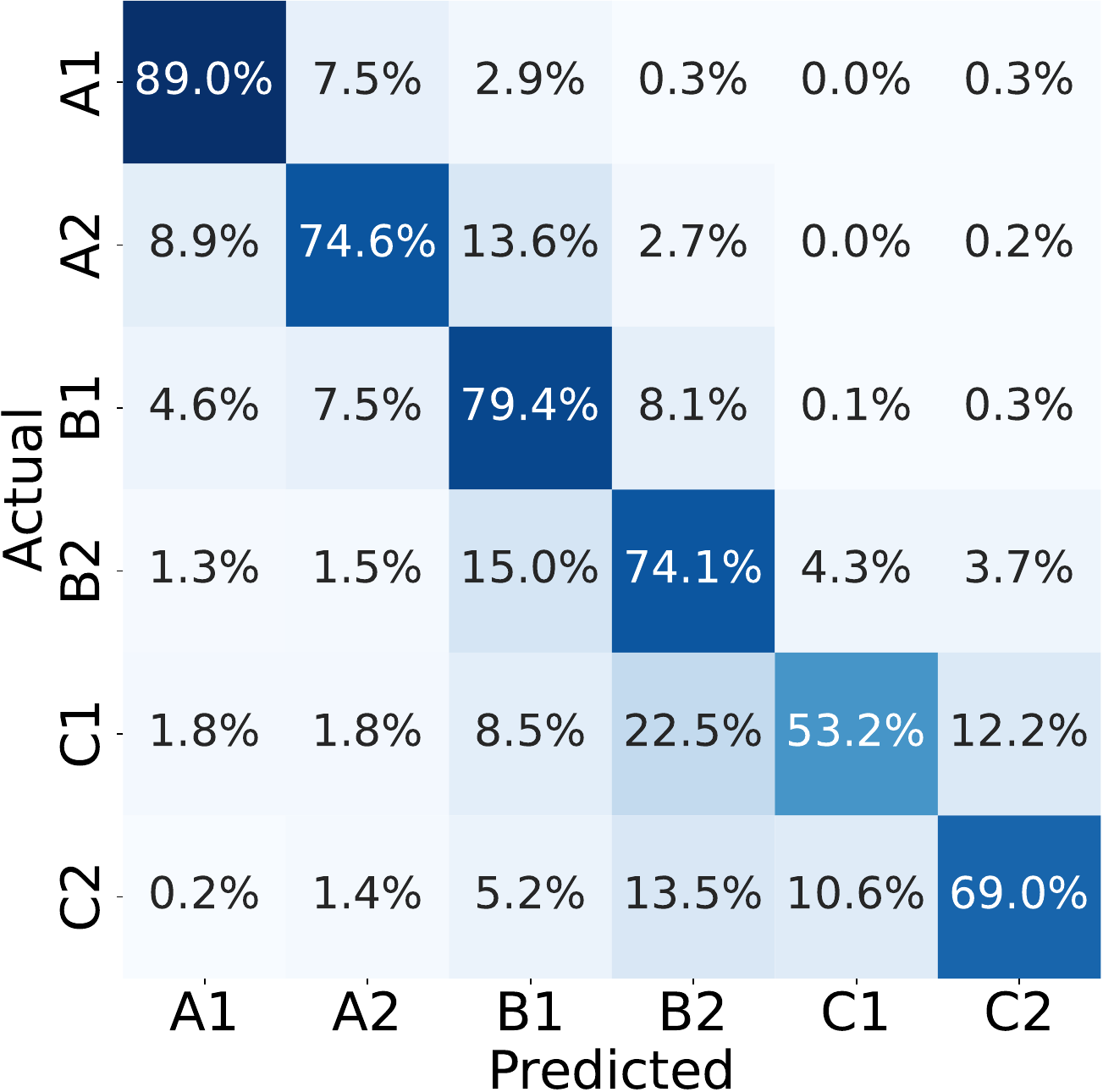}
    \subcaption{FT using Mixture.}\label{fig:ft_evp+semcor}
    \end{minipage}\\\\
    \begin{minipage}[b]{0.33 \linewidth}
    \centering
    \includegraphics[keepaspectratio, scale=0.22]{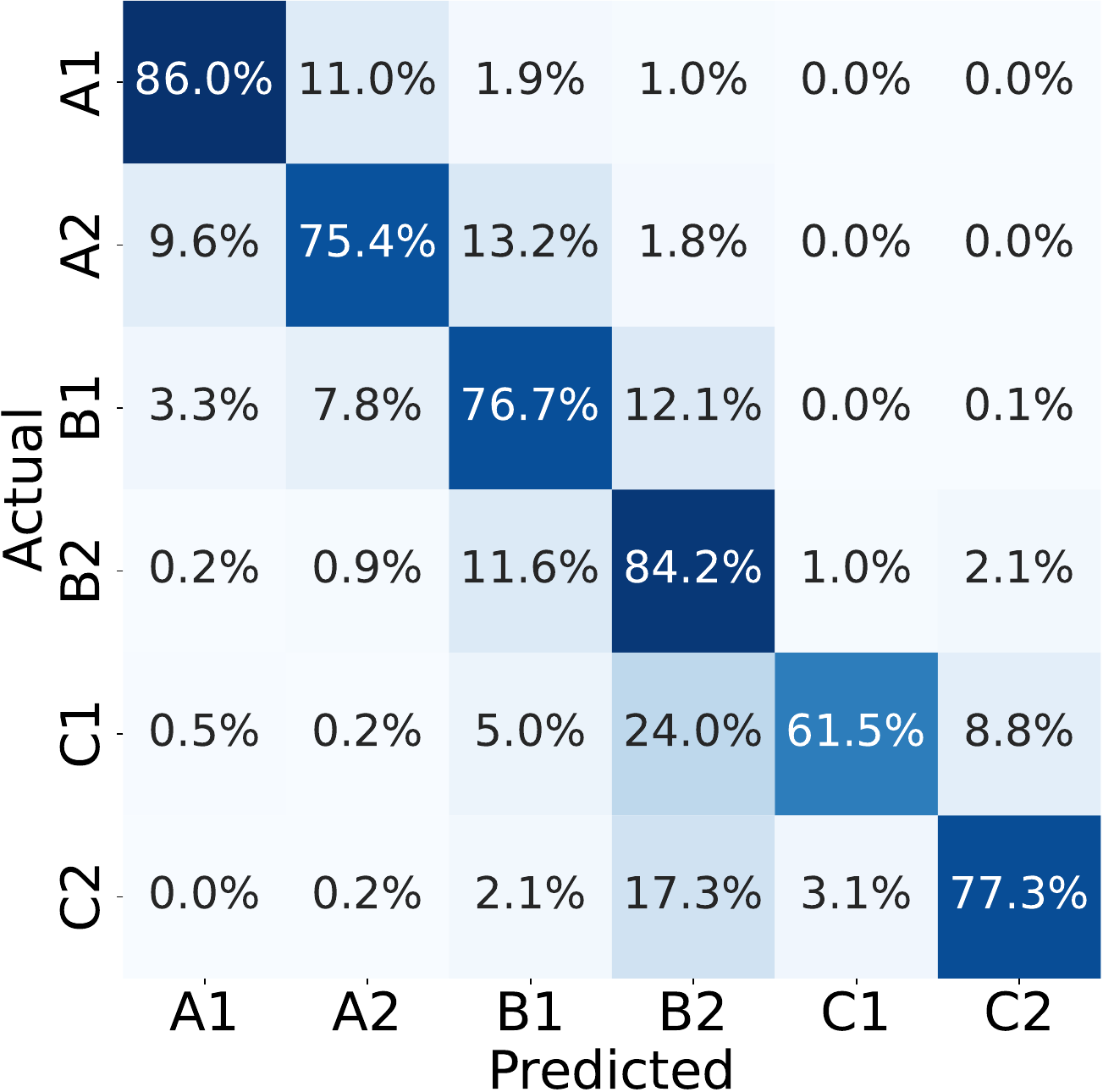}
    \subcaption{FT+KB using EVP.}\label{fig:ft+kb_evp}
    \end{minipage}
    \begin{minipage}[b]{0.33 \linewidth}
    \centering
    \includegraphics[keepaspectratio, scale=0.22]{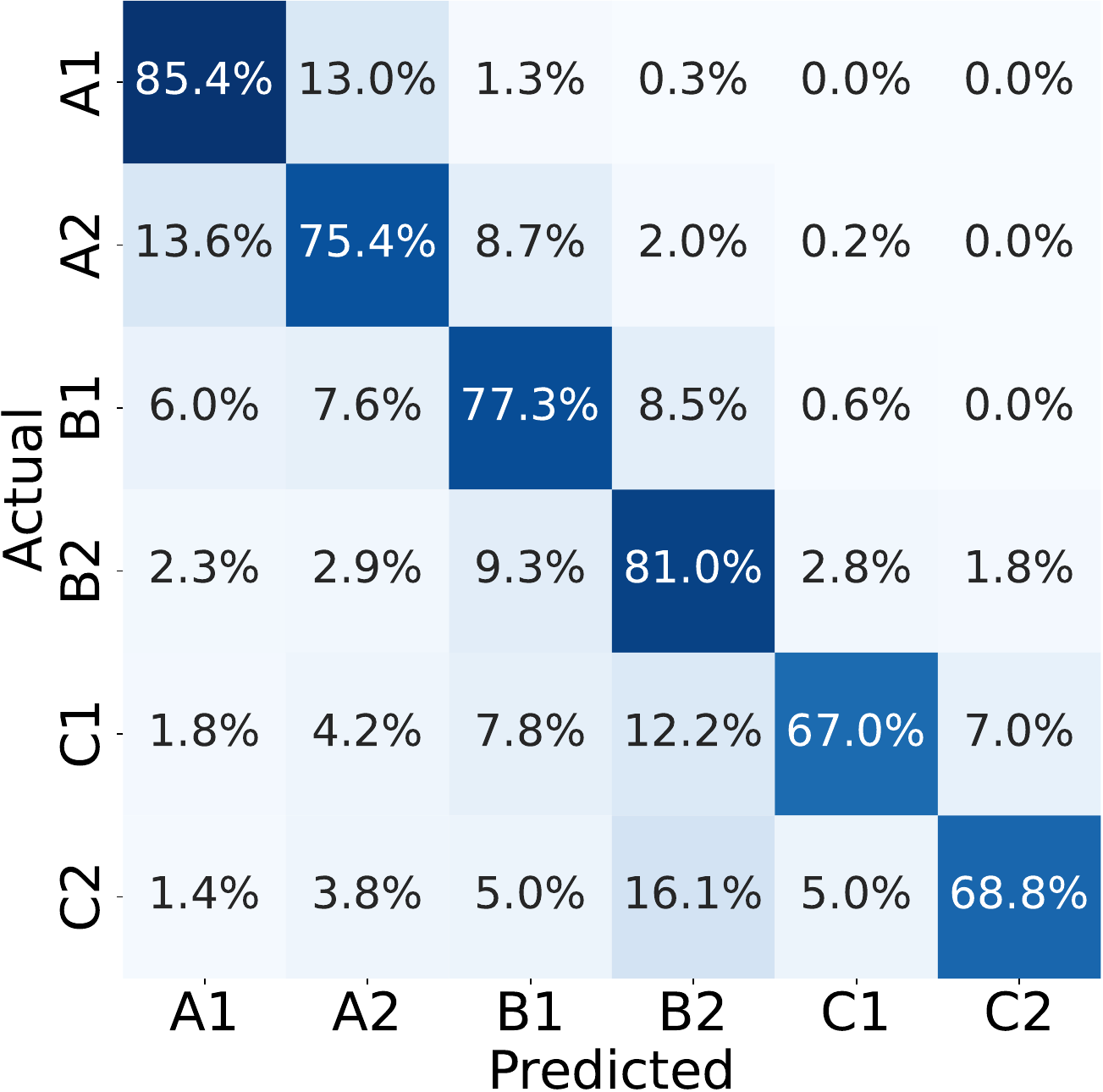}
    \subcaption{FT+KB using SemCor-CEFR.}\label{fig:ft+kb_semcor}
    \end{minipage}
    \begin{minipage}[b]{0.33 \linewidth}
    \centering
    \includegraphics[keepaspectratio, scale=0.22]{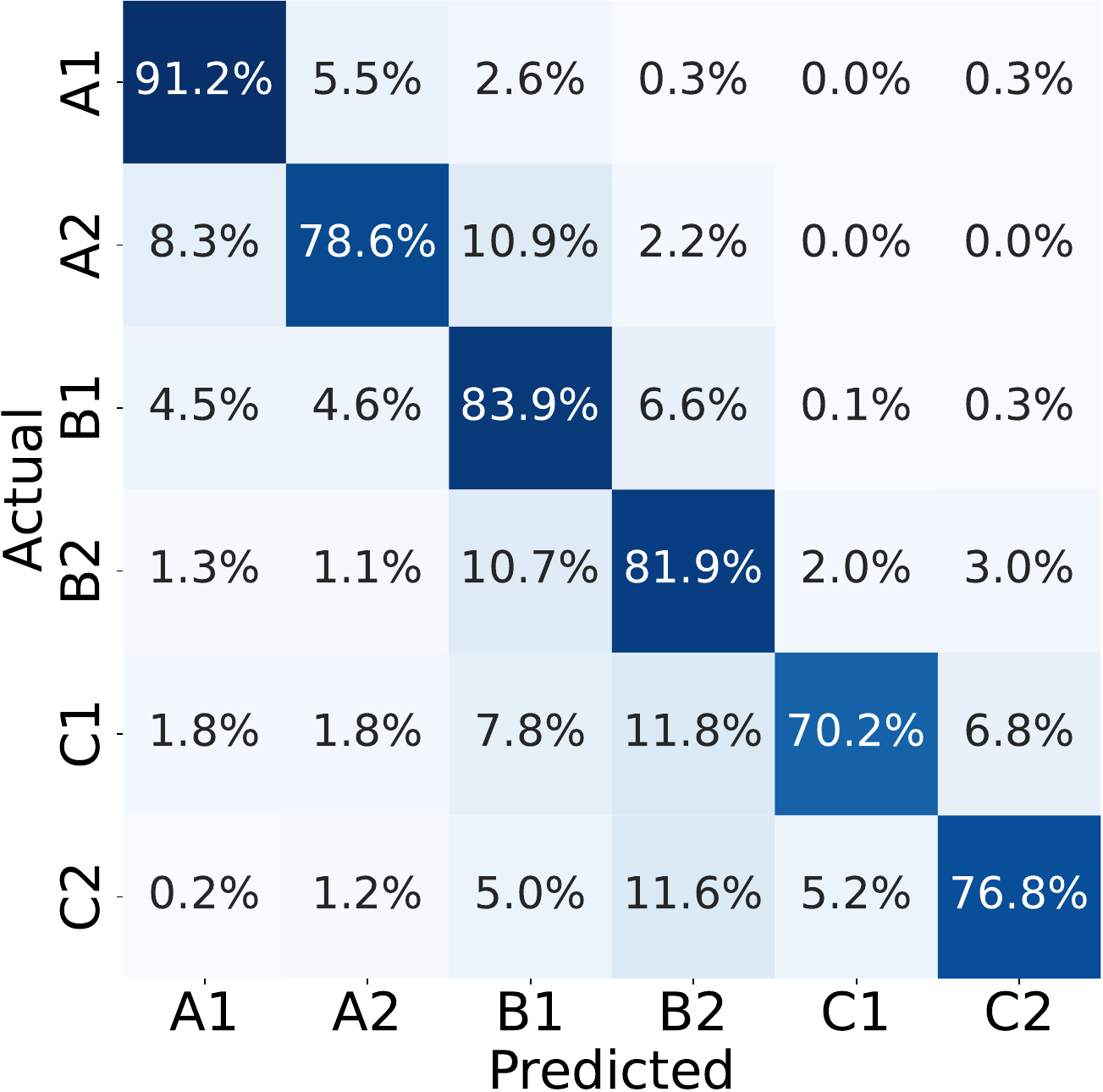}
    \subcaption{FT+KB using Mixture.}\label{fig:ft+kb_evp+semcor}
    \end{minipage}
    \caption{Confusion matrices for each classifier.}\label{fig:matrices}
\end{figure*}
\subsection{Confusion Matrices}
\label{sec:appendix_b}

Figure~\ref{fig:matrices} presents the confusion matrices for each classifier. 
Each matrix element represents the classification probability, calculated as
\begin{displaymath} 
p_{\ell,\widehat{\ell}} = \frac{n_{\ell}(\widehat{\ell})}{n_{\ell}},
\end{displaymath} 
where $n_{\ell}$ denotes the number of target words with the actual CEFR level $\ell$ and $n_{\ell}(\widehat{\ell})$ is the number of those words classified as level $\widehat{\ell}$, i.e., $n_{\ell}=\sum_{\widehat{\ell}}n_{\ell}(\widehat{\ell})$.
The diagonal elements represent the recall for each level; higher values along the diagonal therefore indicate greater accuracy.
Because the CEFR levels are ordinal, misclassifications that fall near the diagonal, that is, those assigned to adjacent levels, are less disruptive for language learners.

The ME6 Contextual models achieve high recall at the lower levels, A1 and A2, as well as at the highest level, C2. 
However, as shown in Figures~\ref{fig:me6_evp} and~\ref{fig:me6_semcor}, when the model is trained on either the EVP or SemCor-CEFR corpus alone, errors at the intermediate and advanced levels, B1 to C2, are more widely distributed.
By contrast, combining both resources (Figure~\ref{fig:me6_evp+semcor}) reduces this dispersion, with most misclassifications occurring between adjacent levels. This result highlights the advantage of jointly leveraging both resources.

As shown in Figure~\ref{fig:zero-shot}, the zero-shot LLM achieves high recall for A1 (82.8\%) and moderate recall for A2 (58.5\%). Performance declines at B1 (36.9\%) and is particularly poor at C1 and C2 (26.0\% and 12.1\%, respectively).
The model frequently misclassifies advanced-level senses as B2, for example 54.2\% of C1 and 44.2\% of C2 cases, indicating a tendency to collapse more difficult senses into an intermediate level. 
Few-shot prompting (Figures~\ref{fig:6-shot} and~\ref{fig:18-shot}) partially alleviates this issue by improving recall at C1 and C2.  
However, recall at A2 declines relative to the zero-shot baseline, and performance at B1 remains similar. These results indicate that the gains from few-shot prompting are uneven across proficiency levels.

By contrast, the FT and FT+KB models substantially improve performance across all CEFR levels. 
When fine-tuned on the Mixture dataset (Figures~\ref{fig:ft_evp+semcor} and~\ref{fig:ft+kb_evp+semcor}), the FT model achieves recall above 70\% for levels A1--B2 and just below 70\% for C2. The FT+KB model further improves recall, exceeding 80\% for B1--B2, reaching approximately 91\% for A1, remaining in the high-70\% range for A2 and C2, and around 70\% for C1.
The corresponding confusion matrices show that errors are concentrated near the diagonal, meaning that most misclassifications occur between adjacent CEFR levels. This pattern reduces potential pedagogical disruption.
Despite these improvements, C1 remains challenging, and C2 instances are often misclassified as B2. 
This pattern also appears when fine-tuning on the EVP or SemCor-CEFR corpora individually (Figures~\ref{fig:ft_evp} and~\ref{fig:ft_semcor}), suggesting that it is not an artifact of the annotation method. Rather, it likely reflects the CEFR-level distribution in the EVP data and characteristics of the fine-tuning process. Further investigation is needed to address these residual errors.

\end{document}